\documentclass{article} 
\usepackage{iclr2027_conference,times}

\usepackage[utf8]{inputenc} 
\usepackage[T1]{fontenc}    
\usepackage[table]{xcolor}
\usepackage{colortbl}
\definecolor{citecolor}{rgb}{0,0,0.5}
\usepackage[pagebackref,colorlinks=true,citecolor=citecolor]{hyperref}

\usepackage{url}            
\usepackage{booktabs}       
\usepackage{amsmath}        
\usepackage{amsfonts}       
\usepackage{nicefrac}       
\usepackage{microtype}      
\usepackage{pgfplots}
\usepackage{pgfplotstable}
\usepackage{mdwlist}
\usepackage{soul}
\usepackage{subcaption}
\usepackage{multirow}
\usepackage{csvsimple}
\usepackage{algorithm}
\usepackage{algpseudocode}

\usepackage[capitalize,noabbrev]{cleveref}
\crefname{equation}{Eq.}{Eqs.}
\crefname{section}{Sec.}{Sec.}
\crefname{figure}{Fig.}{Fig.}
\crefname{table}{Table}{Tables}

\newcommand{\E}{\mathbb{E}}
\newcommand{\KL}{\mathrm{KL}}
\newcommand{\ESS}{\mathrm{ESS}}
\newcommand{\sg}{\mathbf{sg}}
\newcommand{\TokAvg}{\widehat{\mathbb{E}}}

\definecolor{fixedhp}{RGB}{220,20,20}
\definecolor{dataadaptive}{RGB}{0,80,220}
\definecolor{auxchoice}{RGB}{170,30,180}
\definecolor{cb-blue}{RGB}{0, 114, 178}
\definecolor{cb-orange}{RGB}{230, 159, 0}
\definecolor{cb-red}{RGB}{217, 84, 77}
\definecolor{cb-green}{RGB}{0, 158, 115}
\newcommand{\hp}[1]{\textcolor{fixedhp}{\boldsymbol{#1}}}
\newcommand{\hpl}[1]{\textcolor{cb-green}{\boldsymbol{#1}}}
\newcommand{\hph}[1]{\textcolor{fixedhp}{\boldsymbol{#1}}}
\newcommand{\adapt}[1]{\textcolor{dataadaptive}{\boldsymbol{#1}}}
\newcommand{\aux}[1]{\textcolor{auxchoice}{\boldsymbol{#1}}}
\usepgfplotslibrary{fillbetween}
\pgfplotsset{compat=1.18}

\title{Let the Batch Decide, On- or Off-Policy: \\ Effective Sample Size for RL Post-Training}

\author{%
Rasool Fakoor\thanks{Corresponding author: \texttt{rasool.fakoor@gmail.com}}\phantom{$^*$}, Murdock Aubry, Nicholas Stranges, Alexander J. Smola \\
  Boson AI
}

\iclrfinalcopy 
\begin{document}

\maketitle

\begin{abstract}
Reinforcement learning (RL) is structurally harder than supervised learning because the policy changes the data distribution it learns from. The resulting fragility is especially visible in large-model training using nominally on-policy RL, where the training and rollout systems can differ in numerical precision, sampling and other implementation details. The common approach handles this difference by constraining the update with a fixed clip range, so the algorithm becomes more sensitive to its configuration and must be retuned whenever the task, model scale or distribution mismatch changes. We argue that the fragility traces to two linked concerns that this one fixed range has to cover. The first is a trust-region concern: an update should not move the policy too far from its current value. The second is an off-policy concern: data from older or different behavior policies should influence the update only to the extent that the update remains reliable. Neither concern is a constant to set in advance, and the policy ratios of the current batch show how severe each one is. We adopt P3O, a simple yet effective batch-adaptive objective from prior work that replaces fixed clipping with the normalized effective sample size of the per-token policy ratios, which we take against the rollout engine's log-probabilities. The same statistic caps the score-function weight and sets the strength of an off-policy regularizer, so the update stays close to the on-policy update when the ratios are close to one. When stale or mismatched data concentrate the ratios, the update tightens but keeps a nonzero score-function weight on every token. Experiments on math reasoning show that P3O compares favorably with tuned baselines and, under FP8 rollouts on Qwen3-8B-Base, avoids the late collapse of GRPO. P3O adds no new objective hyper-parameters and removes the clip range\footnote{Code is available at \url{https://github.com/FeynRL-project/FeynRL}.}. 

\end{abstract}

\section{Introduction}
\label{sec:introduction}

Post-training large (language) models often means optimizing a policy by sampling completions and scoring them with a reward model, a human-preference-derived objective, or a task verifier~\citep{ziegler2019fine,stiennon2020learning,ouyang2022training,deepseekai2025deepseekr1}. Because these scores are assigned to discrete sampled text and are usually not differentiable through the policy, policy-gradient reinforcement learning (RL) is a natural tool for increasing expected score. RL is a fragile optimization process whose outcomes depend heavily on hyper-parameters and implementation details~\citep{henderson2018deep,engstrom2020implementation,andrychowicz2021what,fakoor2020ddpg,shengyi2022the37implementation}. This fragility sharpens at large-model scale, where runs are expensive, rollouts are costly to regenerate, and small choices (clip range, allowed rollout staleness, etc.) determine whether training improves, stalls, or becomes unstable. These choices are often retuned for each task, model scale, and rollout infrastructure.

Two concerns underlie this fragility. The first is a \emph{trust-region} concern, that each gradient update should not move the policy too far from its current value~\citep{schulman2015trust,schulman2017proximal}. The second is an \emph{off-policy} concern, that data collected by older or different behavior policies should influence the current update only to the extent that the update remains reliable~\citep{Fakoor2020mql,fakoor2024timevarying,espeholt2018impala}. The two concerns are linked, since a surrogate objective estimated from behavior-policy samples is accurate only for policies close to the behavior policy.

Off-policy data are common in practice and arise from several sources, including rollouts reused across optimizer steps, rollouts from earlier policies in asynchronous pipelines~\citep{noukhovitch2024asynchronous}, demonstrations or rollouts from heterogeneous policies (the batch-RL setting;~\citealp{levineOfflineReinforcementLearning2020, fakoor2021continuous, bugdetYao23}), and practical mismatches between the rollout and training engines such as numerical precision~\citep{qi2025fp16}, sampling strategy, or expert routing~\citep{ma2025router}. Mishandling such data is consequential, since stale or mismatched samples can destabilize training, yet discarding them forfeits sample efficiency because each rollout is then consumed by a small number of updates~\citep{fakoorp3o}.

Proximal policy optimization (PPO) and its critic-free variant GRPO cover both concerns with a single fixed-clip mechanism~\citep{schulman2017proximal,shao2024deepseekmath}, where a fixed range controls the per-step policy change and is also relied on to limit the off-policy variance. This is simple, but the range is a strong algorithmic choice made before training starts, and it determines which tokens receive gradient, how much stale data can be reused, and how quickly the policy is allowed to move. Decoupled-loss methods separate the proximal policy used for clipping from the behavior policy used for importance correction and improve learning from stale data~\citep{hilton2021batch}. Recent objectives make the range asymmetric and add dynamic sampling (DAPO;~\citealp{yu2025dapo}), clip a sequence-level ratio (GSPO;~\citealp{zheng2025gspo}), or remove the lower bound and cap a stop-gradient importance weight at a fixed upper value (CISPO;~\citealp{minimax2025m1}).

The premise of this paper is that much of this fragility comes from fixing the clip before training, and that changing the shape of the clip (asymmetric, sequence-level, one-sided) still leaves a range or cap to fix in advance. We build on the observation that the per-token policy ratios of the current batch show how severe each concern is. Using only their normalized effective sample size (ESS;~\citealp{kong1992}), P3O~\citep{fakoorp3o} sets the cap on the score-function weight and regularizes toward the behavior policy, with no new objective hyper-parameter. Unlike fixed-clip methods, whose range is set before training starts, P3O thus adapts the cap and the regularization to how far apart the behavior and current policies are in each batch. When the two match, every ratio is one and, with group-relative advantages, P3O's gradient equals that of GRPO; as they drift apart, the cap tightens and the regularizer grows. The same objective therefore applies whether the batch is on- or off-policy, while fixed-clip objectives may need a decoupled loss~\citep{hilton2021batch} or a retuned clip.

Our main contribution is to adopt P3O for RL post-training of large models, where, to our knowledge, it has not been studied. The original P3O, designed for actor-critic training on games and continuous control, caps only replayed data, since its fresh rollouts are on-policy. In large-model post-training, there can be a log-probability gap between the rollout and training engines even for a fresh rollout, by design (e.g., lower-precision rollouts) or unintentionally (e.g., different kernels or expert routing). We therefore cap every token, with each ratio taken against the rollout engine's recorded log-probability, so one batch can mix several policy versions. We use group-relative advantages in place of a critic. Used unchanged in synchronous and asynchronous training (\cref{fig:async_training}), with the GRPO hyper-parameters of each experiment and no clip sweep, this one objective compares favorably with tuned baselines on math reasoning across several Qwen models. The clearest difference appears when Qwen3-8B-Base is trained in BF16 with FP8 rollouts, where the quantization mismatch reaches every token even though no rollout is stale. By iteration 30, GRPO falls below the untrained model on all five benchmarks, while P3O stays above it (\cref{fig:pass_at_k_fp8_avg,tab:benchmark_results}).

\section{Background}
\label{sec:lm_setting}

Given a pre-trained model, our objective is to fine-tune it using reinforcement learning so that completions sampled from it receive high scores. For a prompt $x$, the model assigns a probability to a completion $y=(y_1,\ldots,y_T)$ autoregressively,
\begin{align}
    \pi_\theta(y \mid x)
    = \prod_{t=1}^{T} \pi_\theta(y_t \mid x,\, y_{<t}),
    \label{eq:autoregressive}
\end{align}
where each $y_t$ is a token from a finite vocabulary $\mathcal{V}$, $y_{<t}=(y_1,\ldots,y_{t-1})$ is the prefix at position $t$, and $T$ is the completion length. To simplify, we write $c_{<t}=(x,y_{<t})$ for the conditioning context.

\paragraph{Token-level MDP.}
We cast this autoregressive generation as a finite-horizon MDP in which the state at position $t$ is $s_t=(x,y_{<t})$, the action is the next token $y_t\in\mathcal{V}$, and a scalar terminal reward $r(x,y)\in\mathbb{R}$ is assigned to the full completion.\footnote{The formulation extends to per-token rewards $r_t(x,y_{\le t})$ by replacing $r(x,y)$ with the reward-to-go $\sum_{t'=t}^{T} r_{t'}(x,y_{\le t'})$ inside the sum in~\cref{eq:pg_grad}.} The agent's objective is to maximize the expected return $J(\pi_\theta) = \E_{x,\, y\sim\pi_\theta(\cdot\mid x)}\!\left[\,r(x,y)\,\right]$, where $x$ is a prompt from the training set and $y$ is a completion sampled from $\pi_\theta$. Using the likelihood-ratio trick~\citep{williams1992simple}, the policy gradient of $J(\pi_\theta)$ is
\begin{align}
    \nabla_\theta J(\pi_\theta)
    =
    \E_{x,\, y\sim\pi_\theta(\cdot\mid x)}\!\left[\,\sum_{t=1}^{T} r(x,y)\,\nabla_\theta\log \pi_\theta(y_t\mid x,y_{<t})\,\right].
    \label{eq:pg_grad}
\end{align}
This is known as the REINFORCE policy-gradient estimator~\citep{williams1992simple}. It is an on-policy estimator: it optimizes the same policy that was used to collect the data. Despite its simplicity, the policy-gradient estimator has high variance because the update uses sampled trajectories and their corresponding rewards as noisy estimates of how good each action was, so the same policy can produce very different rewards across rollouts, especially with long horizons and delayed rewards~\citep{sutton2018reinforcement}. The standard remedy is to subtract a control variate (baseline) $b$ that does not depend on the sampled action $y_t$, replacing $r$ by an advantage $A=r-b$. For any such baseline the gradient remains unbiased, while for a well-chosen $b$ its variance is typically reduced substantially. Actor-critic methods learn a state-dependent value function $V_\phi(s_t)=V_\phi(x,y_{<t})$ and use it as the baseline at each step.

\paragraph{Group-relative advantage.}
Learning a value function alongside the policy is usually expensive when training large models, because the value function can itself be a network of comparable size to the policy and inherits all the difficulties of training a large model. GRPO~\citep{shao2024deepseekmath} sidesteps this by replacing the learned baseline with a within-batch reward statistic. For each prompt $x_p$, it samples a \emph{group} of $G$ completions $\{y_{p,j}\}_{j=1}^{G}$ and uses the group's reward statistics as the baseline. Writing $r_{p,j}:=r(x_p,y_{p,j})$ for the reward of completion $j$ in group $p$, the group-relative advantage is
\begin{align}
    A_{p,j}
    =
    \frac{r_{p,j}-\frac{1}{G}\sum_{k=1}^{G}r_{p,k}}
    {\sqrt{\frac{1}{G-1}\sum_{k=1}^{G}\!\left(r_{p,k}-\frac{1}{G}\sum_{l=1}^{G}r_{p,l}\right)^{\!2}}+\delta},
    \label{eq:grpo_advantage}
\end{align}
where $\delta>0$ avoids division by zero. The group standard deviation rescales the advantage so it is comparable across prompts. Both statistics depend on $r_{p,j}$, so the gradient is biased in general.

\paragraph{Trust region policy optimization.}
A separate optimization concern, beyond the variance and baseline issues above, is that a single gradient update can move $\pi_\theta$ too far from its current value and destabilize training, even when the data are on-policy. Trust-region methods address this by constraining each update to stay close to the current policy. Where TRPO solves this problem with a complex second-order method~\citep{schulman2015trust}, PPO~\citep{schulman2017proximal} takes a simpler approach by using first-order methods to keep new policies close to old. Concretely, PPO samples data from a snapshot $\pi_{\theta_{\mathrm{old}}}$ of the policy, takes several optimizer steps on these data, and clips the per-token policy ratio $\rho_t(\theta)$ before applying it,
\begin{align}
    \mathcal{L}_{\mathrm{PPO}}(\theta)
    = -\E_{\pi_{\theta_{\mathrm{old}}}}\!\left[\min\!\bigl(\rho_t(\theta)\,A,\ \mathrm{clip}(\rho_t(\theta),\,1-\epsilon_\ell,\,1+\epsilon_h)\,A\bigr)\right],\quad \rho_t(\theta) = \frac{\pi_\theta(y_t\mid x,y_{<t})}{\pi_{\theta_{\mathrm{old}}}(y_t\mid x,y_{<t})}
    \label{eq:ppo}
\end{align}
with the clip range $[1-\epsilon_\ell,\,1+\epsilon_h]$ fixed before training. Because $\pi_{\theta_{\mathrm{old}}}$ is close to $\pi_\theta$ in PPO's intended setting, $\rho_t$ stays close to one for most tokens and the clip discourages large policy changes.

\paragraph{From on-policy to off-policy data.}
The on-policy estimator in~\cref{eq:pg_grad} is unbiased but sample-inefficient, since each rollout is consumed by a single gradient step. In practice the data we train on are rarely drawn from $\pi_\theta$ exactly. In such cases the data were sampled from a \emph{behavior policy} $\pi_{\mathrm{b}}$ but the loss is evaluated under the current $\pi_\theta$, so the empirical gradient computed on samples from $\pi_{\mathrm{b}}$ is a biased estimate of $\nabla_\theta J(\pi_\theta)$.

The standard fix is importance sampling~\citep{Fakoor2020mql,fakoor2024timevarying,resnick2013probability}, which restores unbiasedness by changing the measure from $\pi_\theta$ to $\pi_{\mathrm{b}}$. For any function $f$,
\begin{align}
    \E_{z \sim \pi_\theta}[f(z)]
    = \int \pi_\theta(z)\, f(z)\, dz
    = \int \pi_{\mathrm{b}}(z)\, \frac{\pi_\theta(z)}{\pi_{\mathrm{b}}(z)}\, f(z)\, dz
    = \E_{z \sim \pi_{\mathrm{b}}}\!\left[\frac{\pi_\theta(z)}{\pi_{\mathrm{b}}(z)}\, f(z)\right],
    \label{eq:is_change_of_measure}
\end{align}
provided that the Radon-Nikodym derivative $d\pi_\theta/d\pi_{\mathrm{b}}$ is well defined. Applied to the RL objective, this rewrites the expected return as an expectation under the behavior policy,
\begin{align}
    J(\pi_\theta)
    = \E_{x,\, y\sim\pi_\theta(\cdot\mid x)}\!\Big[r(x,y)\Big]
    = \E_{x,\, y\sim\pi_{\mathrm{b}}(\cdot\mid x)}\!\left[\frac{\pi_\theta(y\mid x)}{\pi_{\mathrm{b}}(y\mid x)}\, r(x,y)\right],
    \label{eq:is_objective}
\end{align}
so $J(\pi_\theta)$ can be evaluated on data drawn from $\pi_{\mathrm{b}}$. In our token-level setting we work with the per-token importance ratio
\begin{align}
    \rho_t(\theta)
    =
    \frac{\pi_\theta(y_t\mid x,y_{<t})}
         {\pi_{\mathrm{b}}(y_t\mid x,y_{<t})},
    \label{eq:token_ratio}
\end{align}
which generalizes the PPO-snapshot ratio in~\cref{eq:ppo} to any behavior policy $\pi_{\mathrm{b}}$, and apply this correction directly inside the per-token gradient. Unlike~\cref{eq:is_objective}, this correction weights each token by its own ratio only, so the resulting gradient is biased, as in PPO.

\paragraph{Remark (Importance sampling is unbiased but high variance).}
Importance sampling restores unbiasedness in~\cref{eq:is_objective}, but the per-token ratios $\rho_t(\theta)$ are bounded below by zero and unbounded above, so a few tokens that were unlikely under $\pi_{\mathrm{b}}$ can produce ratios many orders of magnitude larger than one and dominate the gradient estimate. The standard remedy is to clip $\rho_t(\theta)$ before applying it, but as we discuss in \cref{sec:approach}, fixed clipping is itself fragile.

\paragraph{RL in language modeling.}
When fine-tuning language models with RL, the objective is often regularized by a $\KL$ penalty toward a frozen reference policy $\pi_{\theta_0}$, the same policy used to initialize $\pi_\theta$ (typically a pre-trained or supervised-fine-tuned model, the latter trained on ground-truth tokens via teacher forcing;~\citealp{cen2025bridging}),\footnote{More generally, $\pi_{\theta_0}$ can be any reference policy that shares the token vocabulary $\mathcal{V}$ with $\pi_\theta$. When the vocabularies differ, the per-token $\KL$ is no longer well defined and the regularization must be applied at the sequence level.}
\begin{align}
    J_{\mathrm{LM}}(\pi_\theta)
    =
    \E_{x,\, y\sim\pi_\theta(\cdot\mid x)}\!\left[r(x,y)
    -
    \eta \sum_{t=1}^{T} \KL\!\bigl(\pi_\theta(\cdot\mid x, y_{<t})\,\big\|\,\pi_{\theta_0}(\cdot\mid x, y_{<t})\bigr)\right],
    \label{eq:lm_rl_objective}
\end{align}
where $\eta \geq 0$ controls how far $\pi_\theta$ is allowed to drift from $\pi_{\theta_0}$~\citep{ziegler2019fine,stiennon2020learning,ouyang2022training}. The $\KL$ anchor is intended to keep $\pi_\theta$ from drifting too far from $\pi_{\theta_0}$ and potentially from forgetting its language-modeling capabilities. We set $\eta=0$ in all experiments.

\section{Policy-on Policy-off Policy Optimization (P3O)}
\label{sec:approach}

The PPO clipped surrogate in~\cref{eq:ppo} contains two hyper-parameters, $\epsilon_\ell$ and $\epsilon_h$, that specify the clip range $[1-\epsilon_\ell,\,1+\epsilon_h]$ around the per-token ratio $\rho_t$. From here on, $\rho_t$ is the ratio of~\cref{eq:token_ratio}, where the behavior policy $\pi_{\mathrm{b}}$ is the policy whose token log-probabilities the rollout engine recorded (for PPO, $\pi_{\mathrm{b}}=\pi_{\theta_{\mathrm{old}}}$). The effect of $\epsilon_\ell$ and $\epsilon_h$ is most easily seen by considering the per-token contribution to~\cref{eq:ppo} under the two signs of $A$. For tokens with $A>0$, this contribution simplifies to
\begin{align}
    -\min\!\left(\rho_t,\;1+\epsilon_h\right)\,A.
    \label{eq:ppo_pos_adv}
\end{align}
The objective is improved by increasing $\pi_\theta(y_t\mid c_{<t})$, which raises $\rho_t$. Once $\rho_t > 1+\epsilon_h$, the contribution hits the bound $-(1+\epsilon_h)\,A$, and the new policy gains no further credit by increasing the probability of the token. For tokens with $A<0$, the contribution becomes
\begin{align}
    -\max\!\left(\rho_t,\;1-\epsilon_\ell\right)\,A,
    \label{eq:ppo_neg_adv}
\end{align}
where the $\max$ replaces the $\min$ because $A$ is negative. The objective is improved by decreasing $\pi_\theta(y_t\mid c_{<t})$ and hence $\rho_t$. Once $\rho_t < 1-\epsilon_\ell$, the contribution hits $-(1-\epsilon_\ell)\,A$, and the new policy gains no further credit by decreasing the probability of the token. In both cases, $(\epsilon_\ell, \epsilon_h)$ specify how far the new policy can move from $\pi_{\mathrm{b}}$ while continuing to improve the surrogate; clipping is a regularizer that removes the incentive to change the policy dramatically from one update to the next.

The effect of $(\epsilon_\ell, \epsilon_h)$ depends on the regime. When the batch is fresh and on-policy, $\rho_t \approx 1$ for most tokens and the clip is rarely active. When the batch is stale or generated by a different policy, $\rho_t$ can be far from one for many tokens, and $(\epsilon_\ell, \epsilon_h)$ then determines what fraction of the batch contributes to the update. If $(\epsilon_\ell, \epsilon_h)$ is too small, many tokens are clipped before they can contribute; if it is too large, high-variance importance-weighted gradients can destabilize training. A range chosen for one regime may therefore fail in the other, and since it is fixed before training, $(\epsilon_\ell, \epsilon_h)$ in practice has to be tuned for the task, model scale, and degree of off-policy mismatch.

\paragraph{Prior work re-parameterizes the clip but does not remove it.}
Most previous works retain at least one of these two hyper-parameters and differ in how they re-parameterize them. GRPO~\citep{shao2024deepseekmath} uses~\cref{eq:ppo} with a symmetric range $\epsilon_\ell=\epsilon_h$. DAPO~\citep{yu2025dapo} relaxes the symmetry, allowing $\epsilon_\ell\neq \epsilon_h$ to handle positive- and negative-advantage updates separately.\footnote{Note that, in addition to its asymmetric clip, DAPO changes how the data and the reward are used: it removes prompt groups in which all completions receive the same reward (dynamic sampling) and penalizes responses that approach the length limit (overlong penalty). Both changes are complementary to the idea of this paper and can be applied to P3O directly; we do not test this combination, as our focus is on the objective itself. We use DAPO's token-level loss normalization for all methods.} GSPO~\citep{zheng2025gspo} moves the clipped ratio from the token level to the sequence level via a geometric mean, but introduces sequence-level analogues $\epsilon_\ell^{\mathrm{seq}}, \epsilon_h^{\mathrm{seq}}$ that still must be chosen in advance. CISPO~\citep{minimax2025m1} effectively keeps only $\epsilon_h$, as a fixed cap $1+\epsilon_h$ on a stop-gradient importance weight. Across these objectives, $(\epsilon_\ell, \epsilon_h)$ survives in some form (see \cref{tab:objective_comparison}).

\paragraph{Effective sample size.}
Tuning $(\epsilon_\ell, \epsilon_h)$ for large models is expensive. The normalized effective sample size (ESS) of the policy ratios in the current batch~\citep{kong1992,fakoorp3o} provides a way to set the cap from the batch. Concretely, the ESS is given by
\begin{align}
    \ESS(\mathcal{B}; \theta) = \frac{\TokAvg_{\mathcal{B}}[\rho_t]^2}{\TokAvg_{\mathcal{B}}[\rho_t^2]} \in \left[\frac{1}{|\mathcal{B}|},\,1\right],
    \label{eq:ess_def}
\end{align}
where $\mathcal{B}$ is the set of valid response tokens in the current micro-batch, $|\mathcal{B}|$ is its size, and $\TokAvg_{\mathcal{B}}$ denotes the empirical average over $\mathcal{B}$. Each data-parallel worker computes the sums of $\rho_t$ and $\rho_t^2$ and its valid-token count, and these are all-reduced before the ESS is formed (\cref{alg:p3o}). The ratios are already computed for the loss, so this adds one small reduction and no forward pass. We use $e_{\mathcal{B}} = \sg(\ESS(\mathcal{B};\theta))$ to denote its detached value (treated as a constant for backpropagation), where $\sg(\cdot)$ is the stop-gradient operator. The ESS is close to one when the ratios are nearly uniform and falls when a few large ratios dominate, as happens with stale or mismatched data. Observe that, at each context, $\rho_t$ has mean one under $\pi_{\mathrm{b}}$ and second moment one plus the chi-square divergence $\chi^2(\pi_\theta\,\|\,\pi_{\mathrm{b}})$, so the ESS falls as the behavior and current policies move apart. It averages over tokens and does not multiply ratios along a response, so response length by itself does not lower it.

\paragraph{P3O for large-model post-training.}
We adopt the P3O objective of~\citet{fakoorp3o}. P3O replaces the fixed clip in~\cref{eq:ppo} with two terms whose strength is set by the batch ESS,
\begin{align}
    \mathcal{L}_{\mathrm{P3O}}(\theta) = \E_{x,\,y\sim\pi_{\mathrm{b}}(\cdot\mid x)}\!\biggl[\sum_{t=1}^{T} \Bigl( &-\sg(\min\{\rho_t,\,e_{\mathcal{B}}\})\,\log\pi_\theta(y_t\mid c_{<t})\,A \nonumber \\
    &+ (1-e_{\mathcal{B}})\,\KL\!\bigl(\pi_\theta(\cdot\mid c_{<t})\,\big\|\,\pi_{\mathrm{b}}(\cdot\mid c_{<t})\bigr) \Bigr)\biggr],
    \label{eq:p3o_intro}
\end{align}
where the per-token KL, written in the same form as~\cref{eq:lm_rl_objective}, vanishes when $\pi_\theta=\pi_{\mathrm{b}}$. In practice we approximate this KL at the sampled token by $\log\rho_t+1/\rho_t-1$, the estimator GRPO uses for its reference KL~\citep{shao2024deepseekmath}, since we store only that token's log-probability under $\pi_{\mathrm{b}}$. Because $y_t$ is drawn from $\pi_{\mathrm{b}}$ and not from $\pi_\theta$, this term matches the KL in expectation only to second order in $\rho_t-1$ (\cref{app:kl_term}).

P3O makes two structural changes to~\cref{eq:ppo}. First, it replaces the fixed clip range $(\epsilon_\ell,\epsilon_h)$ with the data-driven cap $e_{\mathcal{B}}$. Second, it adds a regularizer with coefficient $(1-e_{\mathcal{B}})$ that grows as the behavior and current policies move apart and vanishes only when all ratios in the batch are equal. Both adaptations are driven by a single statistic of the current batch, so~\cref{eq:p3o_intro} adds no objective hyper-parameter. Replacing $e_{\mathcal{B}}$ by the fixed cap $1+\epsilon_h$ and dropping the KL term gives CISPO~\citep{minimax2025m1}, proposed six years after P3O.

\paragraph{Properties of the update.}
The clip in~\cref{eq:ppo} removes a token's gradient once its ratio passes the bound on the side its advantage favors. In contrast, the score-function weight $\min\{\rho_t,\,e_{\mathcal{B}}\}$ is at most one and never zero, so, as in CISPO, the cap never removes a token's gradient. When $e_{\mathcal{B}}<1$, every token with $\rho_t>e_{\mathcal{B}}$, even one with $\rho_t$ close to one, gets the weight $e_{\mathcal{B}}$. If $\rho_t=1$ for every token, then $e_{\mathcal{B}}=1$, every weight is one and the KL weight is zero, so the P3O gradient equals the GRPO gradient. Off-policy, the cap and the KL term add further bias, as in any truncated or clipped surrogate.

\paragraph{Relation to the original P3O.}
\citet{fakoorp3o} proposed P3O for actor-critic training on Atari and MuJoCo, where fresh rollouts come from the policy being trained, so it applies the cap only to replayed data. Its update adds an unweighted policy-gradient term on fresh rollouts to the capped term $\min\{\rho,c\}$ with $c=\ESS$ and an exact KL penalty with weight $1-\ESS$. In large-model pipelines the rollout and training engines can assign different log-probabilities even to a fresh rollout, so we drop the unweighted term and apply the other two to every token, taking each ratio against the rollout engine's recorded log-probability. Tokens from several policy versions, as in asynchronous training, can then share a batch, and the group-relative advantage of~\cref{eq:grpo_advantage} replaces the critic.

\paragraph{Off-policy data in large-model training.}
At large-model scale, a common fix for off-policy data is a decoupled-loss objective~\citep{hilton2021batch,fu2025areal} that clips against a proximal snapshot $\pi_{\mathrm{prox}}$ rather than against $\pi_{\mathrm{b}}$. This stabilizes training but adds choices that are fixed before training, including the construction of $\pi_{\mathrm{prox}}$, its lifetime, and, in practice, a cap $c_w$ on an outer behavior weight. P3O needs none of them. The same $e_{\mathcal{B}}$ that drives the cap on fresh data falls as the behavior and current policies move apart, so we use~\cref{eq:p3o_intro} unchanged in every setting of~\cref{sec:experiments}, including asynchronous training (\cref{fig:async_training}), without any additional hyper-parameter or special code path. \Cref{app:two_anchor} describes and tests a two-anchor extension of P3O, and \cref{app:related_work} covers further related work.
\begin{table}[t]
\caption{\textbf{Comparison of policy objectives.} Here $\rho_t = \pi_\theta(y_t\mid x,y_{<t})/\pi_{\mathrm{b}}(y_t\mid x,y_{<t})$. \textcolor{fixedhp}{\textbf{Red}}/\textcolor{cb-green}{\textbf{green}}: fixed hyper-parameters (high/low). \textcolor{auxchoice}{\textbf{Purple}}: auxiliary choices. \textcolor{dataadaptive}{\textbf{Blue}}: batch-adaptive quantities. Auxiliary entropy and reference-policy KL terms are omitted.}
\label{tab:objective_comparison}
\centering
\small
\setlength{\tabcolsep}{6pt}
\renewcommand{\arraystretch}{1.4}
\begin{tabular}{@{}lp{0.83\textwidth}@{}}
\toprule
\textbf{Method} & \textbf{Policy objective} \\
\midrule
GRPO / DAPO
&
\(
-\min\left(\rho_t A,\,\mathrm{clip}\left(\rho_t,\,1-\hpl{\epsilon_\ell},\,1+\hph{\epsilon_h}\right)A\right)
\)
\\[0.3em]
GSPO
&
\(
-\min\left(S_\theta(y) A,\,\mathrm{clip}\left(S_\theta(y),\,1-\hpl{\epsilon_\ell^{\mathrm{seq}}},\,1+\hph{\epsilon_h^{\mathrm{seq}}}\right)A\right),\quad S_\theta(y)=\exp\!\left(\tfrac{1}{T}\sum_{t=1}^{T}\log\rho_t\right)
\)
\\[0.3em]
Decoupled
&
\(
-\sg\!\left(\mathrm{clip}\!\left(\tfrac{\aux{\pi_{\mathrm{prox}}}}{\pi_{\mathrm{b}}},\,0,\,\hp{c_w}\right)\right)\min\!\left(\tfrac{\pi_\theta}{\aux{\pi_{\mathrm{prox}}}}A,\,\mathrm{clip}\!\left(\tfrac{\pi_\theta}{\aux{\pi_{\mathrm{prox}}}},\,1-\hpl{\epsilon_\ell},\,1+\hph{\epsilon_h}\right)A\right)
\)
\\[0.3em]
CISPO
&
\(
-\sg(\min\{\rho_t,\,1+\hph{\epsilon_h}\})\,\log\pi_\theta(y_t\mid x,y_{<t})\,A
\)
\quad (lower bound effectively removed)
\\[0.3em]
\midrule
\textbf{P3O}
&
\(
-\sg(\min\{\rho_t,\,\adapt{e_{\mathcal{B}}}\})\,\log\pi_\theta(y_t\mid x,y_{<t})\,A \;+\; (1-\adapt{e_{\mathcal{B}}})\,\KL(\pi_\theta\,\|\,\pi_{\mathrm{b}})
\)
\\
\bottomrule
\end{tabular}
\end{table}
\section{Experiments}
\label{sec:experiments}

We evaluate P3O across three axes: (\textit{i})~sensitivity to the clipping hyper-parameters $(\epsilon_\ell, \epsilon_h)$ that P3O eliminates entirely; (\textit{ii})~stability under off-policy data arising from asynchronous training and from practical mismatches in numerical precision and sampling temperature; and (\textit{iii})~downstream benchmark performance on held-out mathematical reasoning tasks. In all settings we compare against GRPO, which shares the same base objective (\cref{sec:lm_setting}; \cref{alg:grpo,alg:p3o}) but relies on a fixed clip range. P3O uses the GRPO hyper-parameters of each experiment family (\cref{tab:training_params}), and \cref{app:base_study} lists the departures in an additional study on Qwen3-4B-Base.

\subsection{Experiment setup}
Experiments use the open-source models Qwen3-4B-Thinking-2507 and Qwen3-8B-Base~\citep{qwen3technicalreport}, the latter for the FP8 experiments, and Qwen2.5-1.5B-Instruct~\citep{qwen2.5}. We train on a 32,251-problem training split of DeepScaleR-Preview, a dataset of 40,315 mathematics problem-answer pairs compiled from AIME (1984--2023), AMC (prior to 2023), Omni-MATH, and Still~\citep{deepscaler2025}. Binary rewards are assigned by matching the model's output against the DeepScaleR-Preview reference; to isolate algorithmic differences between P3O and GRPO, we use no reward shaping or auxiliary bonuses in either method (the DAPO baseline in \cref{app:dapo} keeps its overlong penalty). GRPO uses $\epsilon=0.4$ in the Qwen3 off-policy experiments. Each configuration is run once. The clip, temperature and FP8 runs use 8 NVIDIA H100 GPUs in a distributed stack that separates training and rollout engines and synchronizes policy weights under a shared scheduler. Training, evaluation, hardware, and benchmark details are in \cref{tab:optimizer_params,tab:training_params,tab:evaluation_params,tab:hardware_config} and \cref{app:exp_details,app:benchmark_results}.

\subsection{Effects of hyper-parameters}

\paragraph{Clip range.}

We test the sensitivity to the clip range by sweeping the symmetric clip $\epsilon_\ell = \epsilon_h = \epsilon \in \{0.2, 0.4, 0.6\}$ for GRPO and contrasting with a single P3O run. These runs take one optimizer step per iteration, so no rollout is stale and $\rho_t$ departs from one only through the log-probability gap between the rollout and training engines. As shown in \cref{fig:clipping_factor_variability}, GRPO's reward trajectory varies substantially with $\epsilon$ on Qwen3-4B-Thinking-2507 and little on Qwen2.5-1.5B-Instruct, while P3O remains stable on both. Other clip-based objectives such as DAPO~\citep{yu2025dapo} and GSPO~\citep{zheng2025gspo} also fix a clip range before training; we report DAPO in two settings with one range each (\cref{app:dapo}) and did not run GSPO or CISPO.

\begin{figure}[t]
    \centering

    {\small%
    \tikz[baseline=-0.6ex]{\draw[cb-red,line width=2.2pt](0,0)--(12pt,0);}\;P3O%
    \hspace{1.6em}%
    \tikz[baseline=-0.6ex]{\draw[cb-blue!35,line width=1.6pt](0,0)--(12pt,0);}\;GRPO, $\epsilon=0.2$%
    \hspace{1.6em}%
    \tikz[baseline=-0.6ex]{\draw[cb-blue,line width=1.6pt](0,0)--(12pt,0);}\;GRPO, $\epsilon=0.4$%
    \hspace{1.6em}%
    \tikz[baseline=-0.6ex]{\draw[cb-blue!55!black,line width=1.6pt](0,0)--(12pt,0);}\;GRPO, $\epsilon=0.6$%
    }

    \begin{subfigure}[b]{0.49\textwidth}
        \centering
        \begin{tikzpicture}
    \begin{axis}[
        width=\textwidth,
        height=0.6\textwidth,
        xlabel={\textbf{Training Steps}},
        ylabel={\textbf{Average Reward}},
        xlabel style={font=\small},
        ylabel style={font=\small},
        grid=major,
        grid style={dashed, gray!30},
        axis x line=bottom,
        axis y line=left,
        enlarge x limits=false,
        ymin=0,
        ymax=0.6,
    ]

    \addplot[color=cb-blue!35, line width=1.2pt] table[x=step, y=clip02, col sep=comma] {plot_data/clipping_ratio_plot_qwen34.csv};
    \addplot[color=cb-blue, line width=1.2pt] table[x=step, y=clip04, col sep=comma] {plot_data/clipping_ratio_plot_qwen34.csv};
    \addplot[color=cb-blue!55!black, line width=1.2pt] table[x=step, y=clip06, col sep=comma] {plot_data/clipping_ratio_plot_qwen34.csv};

    \addplot[color=cb-red, line width=1.8pt] table[x=step, y=p3o_reward, col sep=comma] {plot_data/clipping_ratio_plot_qwen34.csv};

    \end{axis}
\end{tikzpicture}
        \caption{Qwen3-4B-Thinking-2507}
    \end{subfigure}
    \begin{subfigure}[b]{0.49\textwidth}
        \centering
        \begin{tikzpicture}
    \begin{axis}[
        width=\textwidth,
        height=0.6\textwidth,
        xlabel={\textbf{Training Steps}},
        xlabel style={font=\small},
        ylabel style={font=\small},
        grid=major,
        grid style={dashed, gray!30},
        axis x line=bottom,
        axis y line=left,
        enlarge x limits=false,
        ymin=0.15,
        ymax=0.3,
    ]

    \addplot[color=cb-blue!35, line width=1.2pt] table[x=step, y=clip02, col sep=comma] {plot_data/clipping_ratio_plot_qwen25.csv};
    \addplot[color=cb-blue, line width=1.2pt] table[x=step, y=clip04, col sep=comma] {plot_data/clipping_ratio_plot_qwen25.csv};
    \addplot[color=cb-blue!55!black, line width=1.2pt] table[x=step, y=clip06, col sep=comma] {plot_data/clipping_ratio_plot_qwen25.csv};

    \addplot[color=cb-red, line width=1.8pt] table[x=step, y=p3o, col sep=comma] {plot_data/clipping_ratio_plot_qwen25.csv};

    \end{axis}
\end{tikzpicture}
        \caption{Qwen2.5-1.5B-Instruct}
    \end{subfigure}

    \caption{\textbf{Training reward of GRPO with each clip range $\epsilon \in \{0.2, 0.4, 0.6\}$ versus P3O run once with no clip hyper-parameter.} Over the last five steps, P3O's mean reward is 0.510 against 0.196, 0.467 and 0.514 for $\epsilon=0.2$, $0.4$ and $0.6$ on Qwen3-4B-Thinking-2507, and 0.263 against 0.241, 0.268 and 0.263 on Qwen2.5-1.5B-Instruct.}
    \label{fig:clipping_factor_variability}
\end{figure}

\subsection{Off-policy data}

In asynchronous training of Qwen3-4B-Thinking-2507 on GSM8K (\citealp{cobbe2021gsm8k}; \cref{fig:async_training}), GRPO with the decoupled loss is ahead in the middle of training, and P3O is ahead at 22 of the last 23 steps. The two experiments below isolate two other sources of off-policy data and compare P3O with GRPO under each.

\paragraph{Temperature of rollouts.}

We sample rollouts of Qwen3-4B-Thinking-2507 at temperature $T=0.6$ and $T=1.2$. The rollout engine records log-probabilities before temperature scaling and the training engine computes them without temperature, so neither $\rho_t$ (\cref{eq:token_ratio}) nor the ESS (\cref{eq:ess_def}) reflects the change in the sampling distribution, and we report this experiment as an empirical comparison.

\begin{figure}[t]
    \centering

    {\small%
    \tikz[baseline=-0.6ex]{\draw[cb-red,line width=2.2pt](0,0)--(12pt,0);}\;P3O%
    \hspace{1.6em}%
    \tikz[baseline=-0.6ex]{\draw[cb-blue,line width=1.6pt](0,0)--(12pt,0);}\;GRPO%
    }

    \begin{subfigure}[b]{0.49\textwidth}
        \centering
        \begin{tikzpicture}
    \begin{axis}[
        width=\textwidth,
        height=0.6\textwidth,
        xlabel={\textbf{Training Steps}},
        xlabel style={font=\small},
        ylabel={\textbf{Average Reward}},
        ylabel style={font=\small},
        grid=major,
        grid style={dashed, gray!30},
        axis x line=bottom,
        axis y line=left,
        enlarge x limits=false,
        ymin=0,
        ymax=0.8,
    ]

    \addplot[color=cb-blue, line width=1.2pt] table[x=step, y=grpo, col sep=comma] {plot_data/temp_0.6_plot_qwen34.csv};
    \addplot[color=cb-red, line width=1.8pt] table[x=step, y=p3o, col sep=comma] {plot_data/temp_0.6_plot_qwen34.csv};

    \end{axis}
\end{tikzpicture}
        \caption{Qwen3-4B-Thinking-2507 (Temperature 0.6)}
    \end{subfigure}
    \begin{subfigure}[b]{0.49\textwidth}
        \centering
        \begin{tikzpicture}
    \begin{axis}[
        width=\textwidth,
        height=0.6\textwidth,
        xlabel={\textbf{Training Steps}},
        xlabel style={font=\small},
        ylabel style={font=\small},
        grid=major,
        grid style={dashed, gray!30},
        axis x line=bottom,
        axis y line=left,
        enlarge x limits=false,
        ymin=0,
        ymax=0.8,
    ]

    \addplot[color=cb-blue, line width=1.2pt] table[x=step, y=grpo, col sep=comma] {plot_data/temp_1.2_plot_qwen34.csv};
    \addplot[color=cb-red, line width=1.8pt] table[x=step, y=p3o, col sep=comma] {plot_data/temp_1.2_plot_qwen34.csv};

    \end{axis}
\end{tikzpicture}
        \caption{Qwen3-4B-Thinking-2507 (Temperature 1.2)}
    \end{subfigure}
\caption{\textbf{Training reward on off-policy data introduced through the varied sampling temperature of rollouts.}
Sampling rollouts at a temperature other than 1.0 shifts the distribution of sampled tokens. At $T=1.2$ only GRPO collapses; at $T=0.6$ both methods do. Qwen2.5-1.5B-Instruct results are in \cref{fig:temp_variability_appendix}.}
\label{fig:temp_variability}
\end{figure}

As shown in \cref{fig:temp_variability}, at $T=1.2$ the two methods stay within 0.04 of each other up to step 19 and peak at nearly the same reward (0.694 for P3O and 0.697 for GRPO). GRPO then falls from 0.567 at step 21 to 0.199 at step 23 and ends at 0.089, while P3O stays between 0.56 and 0.70 from step 10 to step 36. At $T=0.6$ both methods peak at step 13 and then collapse, and P3O is below GRPO at 15 of the 22 steps both runs share, so P3O does not prevent this failure. On Qwen2.5-1.5B-Instruct (\cref{fig:temp_variability_appendix}), both methods also collapse at $T=0.6$, and neither degrades at $T=1.2$.

\paragraph{BF16 train + FP8 rollout.}

The practical off-policy mismatch described in \cref{sec:introduction} is directly instantiated by mixed-precision pipelines: rollouts generated by an FP8-quantized policy carry different token-level log-probabilities than the BF16 training model, shifting the per-token ratio $\rho_t$ (\cref{eq:token_ratio}) away from one. Rollouts generated in lower precision can be faster to generate~\citep{vllm_fp8_2024}. A common training strategy is to use BF16 precision for training
and FP8 while creating rollouts, but this strategy can lead to collapse in reward performance, particularly for long rollouts~\citep{xi2026jetrlenablingonpolicyfp8}.
In practice, dynamic quantization methods are used where the latest trained policy is moved to the rollout engine with BF16 precision, then dynamically quantized to FP8 for rollout generation~\citep{vllm_fp8_2024,xi2026jetrlenablingonpolicyfp8}. We test whether P3O (\cref{eq:p3o_intro}) handles this mismatch without any change to the training configuration. GRPO takes its ratio against the same recorded log-probabilities and uses no decoupled loss or other correction for the mismatch. Both runs are configured for 500 iterations with 50 warmup iterations, so the learning rate is still rising at the evaluated iterations 15 and 30 (\cref{tab:optimizer_params}). As shown in \cref{fig:fp8_training}a, GRPO on Qwen3-8B-Base is ahead during the first 15 steps, and the two methods peak at step 16 at nearly the same reward (0.542 for GRPO and 0.550 for P3O). GRPO then falls below its starting reward from step 22 and below 0.1 from step 27, while P3O stays above 0.49 up to step 35. On Qwen2.5-1.5B-Instruct (\cref{fig:fp8_training}b) the curves overlap and neither method degrades; GRPO is slightly higher over the last five steps (0.248 against 0.234). The gain from P3O is thus the absence of the late collapse on Qwen3-8B-Base, not a higher peak.

\begin{figure}[t]
    \centering
    {\small%
    \tikz[baseline=-0.6ex]{\draw[cb-red,line width=2.2pt](0,0)--(12pt,0);}\;P3O%
    \hspace{1.6em}%
    \tikz[baseline=-0.6ex]{\draw[cb-blue,line width=1.6pt](0,0)--(12pt,0);}\;GRPO%
    }

    \begin{subfigure}[b]{0.49\textwidth}
        \centering
        \begin{tikzpicture}
    \begin{axis}[
        width=\textwidth,
        height=0.6\textwidth,
        xlabel={\textbf{Training Steps}},
        ylabel={\textbf{Average Reward}},
        xlabel style={font=\small},
        ylabel style={font=\small},
        grid=major,
        grid style={dashed, gray!30},
        axis x line=bottom,
        axis y line=left,
        enlarge x limits=false,
        ymin=0,
        ymax=0.6,
    ]

    \addplot[color=cb-blue, line width=1.2pt] table[x=step, y=grpo, col sep=comma] {plot_data/fp8_plot_qwen34.csv};
    \addplot[color=cb-red, line width=1.8pt] table[x=step, y=p3o, col sep=comma] {plot_data/fp8_plot_qwen34.csv};

    \end{axis}
\end{tikzpicture}
        \caption{Qwen3-8B-Base}
    \end{subfigure}
    \begin{subfigure}[b]{0.49\textwidth}
        \centering
        \begin{tikzpicture}
    \begin{axis}[
        width=\textwidth,
        height=0.6\textwidth,
        xlabel={\textbf{Training Steps}},
        xlabel style={font=\small},
        grid=major,
        grid style={dashed, gray!30},
        axis x line=bottom,
        axis y line=left,
        enlarge x limits=false,
        ymin=0.1,
        ymax=0.3,
    ]

    \addplot[color=cb-blue, line width=1.2pt] table[x=step, y=grpo, col sep=comma] {plot_data/fp8_plot_qwen25.csv};
    \addplot[color=cb-red, line width=1.8pt] table[x=step, y=p3o, col sep=comma] {plot_data/fp8_plot_qwen25.csv};

    \end{axis}
\end{tikzpicture}
        \caption{Qwen2.5-1.5B-Instruct}
    \end{subfigure}

    \caption{\textbf{Training reward on off-policy data introduced through the BF16 train + FP8 rollout scheme.}
    Because accuracy can collapse with long FP8 rollouts~\citep{xi2026jetrlenablingonpolicyfp8}, we use 16,384-token rollouts for Qwen3-8B-Base.
    On Qwen3-8B-Base, GRPO collapses after its peak at step 16 while P3O does not.
    On Qwen2.5-1.5B-Instruct, the curves overlap and neither degrades.}
    \label{fig:fp8_training}
\end{figure}

\subsection{Benchmark results}

To test whether the differences in training reward carry over to held-out task performance, we evaluate the Qwen3-4B-Thinking-2507 checkpoints of the clip study and the Qwen3-8B-Base checkpoints of the FP8 study on five mathematical reasoning benchmarks (\cref{tab:benchmark_results} in the appendix). These are AIME24~\citep{aime24}, AIME25~\citep{aime25},
AIME26~\citep{aime26}, AMO-Bench~\citep{an2025amobench}, and AMC~\citep{amc12_2024}. AIME25 consists of the 15 problems of AIME 2025 I, and AMC of 45 problems from the 2024 AMC 12A and 12B. The DeepScaleR-Preview training set contains no samples from these benchmarks, so the table measures generalization.

\Cref{fig:benchmark_pass_at_k} shows that the training-time stability differences seen in the reward curves also matter at evaluation time. In the clip study (\cref{fig:pass_at_k_clip_avg}), P3O is above the average of the three GRPO runs at every $k$, although GRPO with $\epsilon=0.6$ alone is above P3O on every benchmark at a 16,384-token budget (\cref{tab:thinking_16k}). At this budget, every clip-study checkpoint, trained with 1,024-token rollouts, scores below the untrained model on every benchmark (\cref{app:thinking16k}). In a clip study on Qwen3-4B-Base with a DAPO baseline (\cref{tab:base_benchmark}), P3O lies between the best and the worst GRPO clip on every benchmark but is the best on none. In the FP8 rollout setting (\cref{fig:pass_at_k_fp8_avg}), P3O retains benchmark performance from iteration 15 to iteration 30, while GRPO degrades sharply.


\begin{figure}[t]
\centering

{\small%
\tikz[baseline=-0.6ex]{\draw[black,line width=1.6pt](0,0)--(12pt,0);}\;Untrained%
\hspace{1.6em}%
\tikz[baseline=-0.6ex]{\draw[cb-blue,line width=1.6pt](0,0)--(12pt,0);}\;GRPO%
\hspace{1.6em}%
\tikz[baseline=-0.6ex]{\draw[cb-red,line width=2.2pt](0,0)--(12pt,0);}\;P3O%
\hspace{1.6em}%
\tikz[baseline=-0.6ex]{\draw[gray!50!black,line width=1.6pt,dashed](0,0)--(12pt,0);}\;iter~15 (right)%
\hspace{1.6em}%
\tikz[baseline=-0.6ex]{\draw[gray!50!black,line width=1.6pt](0,0)--(12pt,0);}\;iter~30 (right)%
}

\vspace{0.5em}

\begin{subfigure}[b]{0.49\textwidth}
\centering
\begin{tikzpicture}
\begin{axis}[
    width=\textwidth,
    height=0.6\textwidth,
    xlabel={\textbf{$k$}},
    ylabel={\textbf{Pass@$k$}},
    xlabel style={font=\small},
    ylabel style={font=\small},
    xmin=1, xmax=16, xtick={1,4,8,12,16},
    ymin=0,
    grid=major,
    grid style={dashed, gray!30},
    axis x line=bottom,
    axis y line=left,
    enlarge x limits=false,
    tick label style={font=\small},
]

\addplot[black, line width=1.2pt, solid]
    table[x=k, y=baseline_4k, col sep=comma]
    {plot_data/pass_at_k_clip_avg.csv};

\addplot[cb-blue, line width=1.2pt, solid]
    table[x=k, y=grpo_mean, col sep=comma]
    {plot_data/pass_at_k_clip_avg.csv};

\addplot[cb-red, line width=1.8pt, solid]
    table[x=k, y=p3o, col sep=comma]
    {plot_data/pass_at_k_clip_avg.csv};

\end{axis}
\end{tikzpicture}
\caption{Clip variants (4K-token eval)}
\label{fig:pass_at_k_clip_avg}
\end{subfigure}
\hfill
\begin{subfigure}[b]{0.49\textwidth}
\centering
\begin{tikzpicture}
\begin{axis}[
    width=\textwidth,
    height=0.6\textwidth,
    xlabel={\textbf{$k$}},
    ylabel={\textbf{Pass@$k$}},
    xlabel style={font=\small},
    ylabel style={font=\small},
    xmin=1, xmax=16, xtick={1,4,8,12,16},
    ymin=0,
    grid=major,
    grid style={dashed, gray!30},
    axis x line=bottom,
    axis y line=left,
    enlarge x limits=false,
    tick label style={font=\small},
]

\addplot[black, line width=1.2pt, solid]
    table[x=k, y=baseline_16k, col sep=comma]
    {plot_data/pass_at_k_fp8_avg.csv};

\addplot[cb-blue, line width=1.2pt, dashed]
    table[x=k, y=fp8_grpo_iter15, col sep=comma]
    {plot_data/pass_at_k_fp8_avg.csv};
\addplot[cb-blue, line width=1.2pt, solid]
    table[x=k, y=fp8_grpo_iter30, col sep=comma]
    {plot_data/pass_at_k_fp8_avg.csv};

\addplot[cb-red, line width=1.8pt, dashed]
    table[x=k, y=fp8_p3o_iter15, col sep=comma]
    {plot_data/pass_at_k_fp8_avg.csv};
\addplot[cb-red, line width=1.8pt, solid]
    table[x=k, y=fp8_p3o_iter30, col sep=comma]
    {plot_data/pass_at_k_fp8_avg.csv};

\end{axis}
\end{tikzpicture}
\caption{FP8 variants (16K-token eval)}
\label{fig:pass_at_k_fp8_avg}
\end{subfigure}

\caption{\textbf{Pass@$k$ averaged over all five held-out benchmarks
(AIME24/25/26, AMO-Bench, AMC).}
\textit{Left}: clip variants of Qwen3-4B-Thinking-2507 (iter~35) at 4K-token evaluation; GRPO is averaged over
$\epsilon\in\{0.2,0.4,0.6\}$, and Untrained is Qwen3-4B-Thinking-2507 before training.
P3O is above this average at every $k$.
\textit{Right}: BF16 train + FP8 rollout variants of Qwen3-8B-Base at 16K-token evaluation, and Untrained is Qwen3-8B-Base before training.
At iter~30 the average pass@$k$ of GRPO is at most 0.060 and lies below the untrained model at every $k$, while that of P3O is above it.
}
\label{fig:benchmark_pass_at_k}
\end{figure}

\section{Discussion}
\label{sec:discussion}

RL is structurally and algorithmically fragile, and large-model post-training sharpens that fragility because rollouts are expensive, hyper-parameters are consequential, and tuning costs compound with model scale. Recent approaches respond by adding or re-parameterizing fixed choices (asymmetric clip ranges, behavior-weight caps, etc.) tuned per task and model. We take the opposite path. The amount of trust placed in an update should be adaptive rather than pre-committed before training begins. Using the normalized effective sample size of the current policy ratios, the P3O objective replaces fixed clipping with a batch-adaptive score-function cap and matching regularizer, which behaves like the on-policy update when the ratios are close to one and tightens on stale or mismatched data. This does not make arbitrary off-policy data reliable, since meaningful token-level ratios and adequate support in the behavior data are still required (\cref{app:ess_details}), but it turns off-policy mismatch into a measured batch property and removes the need for a clip range or behavior-weight cap in the objective.

Removing the fixed clip also opens space for downstream work. Because P3O makes off-policy data directly usable inside one objective, future methods can combine off-policy reuse (for sample efficiency) with on-policy collection (for exploration) without retuning a clip range or behavior-weight cap for each setting. We view this paper as a starting point for that direction and for batch-adaptive methods that improve large-model post-training by reducing the surface of pre-committed knobs.

\clearpage
\bibliography{references}
\bibliographystyle{iclr2027_conference}


\newpage
\appendix

\suppressfloats[t]

\section{Algorithm pseudocode}
\label{app:algorithms}

We present pseudocode for GRPO and P3O as implemented in our training framework.
Both algorithms share the same rollout phase and use group-relative advantages~(\cref{eq:grpo_advantage}); they differ only in how the policy update is computed.
GRPO clips the per-token policy ratio with a fixed clip range $(\epsilon_\ell, \epsilon_h)$, while P3O replaces this fixed clip with a batch-adaptive ESS cap and an adaptive KL regularizer.
Throughout both algorithms, $\langle \cdot \rangle_M$ denotes the sum over the valid (non-padded) tokens indicated by mask $M$ divided by the number of valid tokens in the whole rollout batch, $\mathbf{sg}(\cdot)$ is the stop-gradient operator, and $c_{<t} = (x, y_{<t})$ is the conditioning context at position $t$.
Thus, within a rollout batch, the loss is the average over completions of the summed per-token terms, multiplied by the number of completions over the number of valid tokens, so this normalization changes the scale of the gradient of each batch but not its direction. GRPO uses the same normalization.
We set $\beta_{\mathrm{ent}}=0$ in all experiments ($H_\theta$ is the entropy of $\pi_\theta(\cdot\mid c_{<t})$), and both algorithms omit the reference-KL term of~\cref{eq:lm_rl_objective}, since $\eta=0$.

\begin{algorithm}[H]
\caption{GRPO}
\label{alg:grpo}
\begin{algorithmic}[1]
\Require Policy $\pi_\theta$; group size $G$; clip $(\epsilon_\ell, \epsilon_h)$; entropy coefficient $\beta_{\mathrm{ent}}$
\Statex
\Statex \textit{Rollout:} for each prompt $x_p$, sample $G$ completions with the rollout engine, record their token log-probabilities $\log\pi_{\mathrm{b}}$, and compute
\[
  A_{p,j} = \frac{r_{p,j} - \bar{r}_p}{\sqrt{\frac{1}{G-1}\sum_k\!\left(r_{p,k}-\bar{r}_p\right)^2}+\delta}
\]
\Statex where $\bar{r}_p=\frac{1}{G}\sum_k r_{p,k}$, and $A_t=A_{p,j}$ for every token $y_t$ of $y_{p,j}$.
\Statex \textit{Update:} for each micro-batch $\mathcal{B}$ with mask $M$:
\State $\rho_t \gets \exp\!\bigl(\log\pi_\theta(y_t\!\mid\!c_{<t}) - \log\pi_{\mathrm{b}}(y_t\!\mid\!c_{<t})\bigr)$
\State $\mathcal{L} \gets -\bigl\langle\min\!\bigl(\rho_t A_t,\;\mathrm{clip}(\rho_t,1{-}\epsilon_\ell,1{+}\epsilon_h)\,A_t\bigr)\bigr\rangle_{\!M} - \beta_{\mathrm{ent}}\langle H_\theta\rangle_{\!M}$
\State Backward pass; accumulate gradients
\Statex \textit{After all micro-batches:} one optimizer step
\end{algorithmic}
\end{algorithm}

\begin{algorithm}[H]
\caption{P3O}
\label{alg:p3o}
\begin{algorithmic}[1]
\Require Policy $\pi_\theta$; group size $G$; entropy coefficient $\beta_{\mathrm{ent}}$ \hfill\textit{(no clip range)}
\Statex
\Statex \textit{Rollout:} identical to \cref{alg:grpo}
\Statex \textit{Update:} for each micro-batch $\mathcal{B}$ with mask $M$:
\State $\rho_t \gets \exp\!\bigl(\log\pi_\theta(y_t\!\mid\!c_{<t}) - \log\pi_{\mathrm{b}}(y_t\!\mid\!c_{<t})\bigr)$
\State $e_{\mathcal{B}} \gets \sg\Bigl(\bigl(\sum_{M}\rho_t\bigr)^{\!2}\big/\bigl(|M|\cdot\sum_{M}\rho_t^2\bigr)\Bigr)$ \hfill (sums and $|M|$ all-reduced across data-parallel workers)
\State $\mathcal{L} \gets -\langle\sg(\min\{\rho_t,e_{\mathcal{B}}\})\log\pi_\theta A_t\rangle_M + (1{-}e_{\mathcal{B}})\langle\log\rho_t+1/\rho_t-1\rangle_M - \beta_{\mathrm{ent}}\langle H_\theta\rangle_M$
\State Backward pass; accumulate gradients
\Statex \textit{After all micro-batches:} one optimizer step
\end{algorithmic}
\end{algorithm}

\paragraph{The KL term.}
\label{app:kl_term}
\Cref{alg:p3o} replaces the per-token KL of~\cref{eq:p3o_intro} by $h(\rho_t)=\log\rho_t+1/\rho_t-1$ at the sampled token, which is nonnegative and zero only at $\rho_t=1$. Since $y_t$ is drawn from $\pi_{\mathrm{b}}$, its mean at each context is given by
\begin{align*}
    \E_{\pi_{\mathrm{b}}}[h(\rho_t)]=\chi^2(\pi_{\mathrm{b}}\,\|\,\pi_\theta)-\KL(\pi_{\mathrm{b}}\,\|\,\pi_\theta),
\end{align*}
and not $\KL(\pi_\theta\,\|\,\pi_{\mathrm{b}})=\E_{\pi_{\mathrm{b}}}[\rho_t h(\rho_t)]$. GRPO uses $h$ for its reference KL with tokens drawn from $\pi_{\theta_{\mathrm{old}}}$, which is close to $\pi_\theta$ in its intended setting, so it uses no such weight. For example, with $\pi_\theta=(0.9,0.1)$ and $\pi_{\mathrm{b}}=(0.5,0.5)$, the KL is about $0.368$, whereas the mean of $h$ is about $1.267$. With $\delta_t=\rho_t-1$, both $h(\rho_t)$ and $\rho_t h(\rho_t)$ equal $\delta_t^2/2+O(\delta_t^3)$, so both means equal $\chi^2(\pi_\theta\,\|\,\pi_{\mathrm{b}})/2$ to second order, and the loss term matches the KL when the two policies are close. \citet{fakoorp3o} write $\KL(\pi_{\mathrm{b}}\,\|\,\pi_\theta)$ in their gradient but state that they use $\KL(\pi_\theta\,\|\,\pi_{\mathrm{b}})$, which they compute from stored action distributions. We keep this direction, but since we store only the sampled token's log-probability, we use $h(\rho_t)$ in its place.
The derivative of $(1-e_{\mathcal{B}})h(\rho_t)$ with respect to $\log\pi_\theta(y_t\mid c_{<t})$ is $(1-e_{\mathcal{B}})(1-1/\rho_t)$, whose magnitude grows without bound as $\rho_t\to0$ when $e_{\mathcal{B}}<1$. Thus the cap bounds the score-function weight but not the whole per-token gradient.

\paragraph{Computing the ESS.}
\label{app:ess_details}
Each micro-batch of~\cref{alg:p3o} gets its own $e_{\mathcal{B}}$, computed over its tokens and those of the micro-batches that the other data-parallel workers process at the same time (\cref{tab:training_params}). Since each $e_{\mathcal{B}}$ depends on which tokens share a micro-batch, the same rollouts split differently can give a different cap and KL weight, and thus a different update. The micro-batches are formed by a seeded random shuffle of the rollout batch.
Note that the empirical ESS measures how concentrated the observed ratios are; it is not a distance between $\pi_\theta$ and $\pi_{\mathrm{b}}$. It does not change when every ratio is multiplied by the same constant: if $\rho_t=0.5$ for every token, then $e_{\mathcal{B}}=1$, the KL weight is zero and every score-function weight is $0.5$. When tokens are sampled from $\pi_{\mathrm{b}}$ and $\pi_{\mathrm{b}}$ is positive wherever $\pi_\theta$ is, $\rho_t$ has mean one at each context, so such a scaling is unlikely in a large batch, and the relation to $\chi^2(\pi_\theta\,\|\,\pi_{\mathrm{b}})$ after~\cref{eq:ess_def} holds in this population sense. However, a finite batch can miss rare tokens with large ratios, and its ESS can then be much higher than the population value. We therefore use the ESS to set the cap and the KL weight, not as a guarantee that the two policies are close.

\section{Benchmark results}
\label{app:benchmark_results}

\Cref{tab:benchmark_results} reports pass@1 for each checkpoint on all five held-out benchmarks. Pass@1 is the fraction of correct completions among 16 samples per problem, averaged over the problems of a benchmark. For $k\le16$, pass@$k$ (\cref{fig:benchmark_pass_at_k}) is the probability that at least one of $k$ samples drawn without replacement from these 16 is correct, averaged in the same way. The untrained Qwen3-8B-Base and the DAPO checkpoints were evaluated in a later round than the FP8 GRPO and P3O checkpoints, the untrained model with a later revision of the evaluation code (the DAPO evaluations do not record their revision). All use the same sampling settings (\cref{tab:evaluation_params}).

\begin{table}[t]
\centering
\footnotesize
\setlength{\tabcolsep}{4pt}
\renewcommand{\arraystretch}{1.2}
\begin{tabular}{lccccc}
\toprule
\textbf{Training Method} & \textbf{AIME24} & \textbf{AIME25} & \textbf{AIME26} & \textbf{AMO-Bench} & \textbf{AMC} \\
\midrule
\rowcolor[gray]{0.95} \multicolumn{6}{l}{\textit{Clip Variants (Qwen3-4B-Thinking-2507, $\epsilon \in \{0.2,0.4,0.6\}$, 4K tokens)}} \\
Untrained model (4K tokens)  & 0.029 & 0.033 & 0.006 & 0.008 & 0.217 \\
GRPO (clip avg) & $0.122^{\pm 0.083}$ & $0.160^{\pm 0.123}$ & $0.126^{\pm 0.087}$ & $\mathbf{0.012}^{\pm 0.007}$ & $0.381^{\pm 0.195}$ \\
P3O & \textbf{0.165} & \textbf{0.183} & \textbf{0.160} & 0.010 & \textbf{0.493} \\
\midrule
\rowcolor[gray]{0.95} \multicolumn{6}{l}{\textit{FP8 Variants (Qwen3-8B-Base, BF16 train + FP8 rollout, 16K tokens)}} \\
Untrained model (16K tokens) & 0.077 & 0.121 & 0.056 & 0.021 & 0.228 \\
FP8 Rollout GRPO Iter 15 & 0.154 & 0.250 & 0.160 & 0.021 & 0.499 \\
FP8 Rollout GRPO Iter 30 & 0.002 & 0.000 & 0.002 & 0.019 & 0.029 \\
FP8 Rollout DAPO Iter 15 & 0.071 & 0.142 & 0.065 & 0.025 & 0.349 \\
FP8 Rollout DAPO Iter 30 & 0.006 & 0.004 & 0.004 & 0.015 & 0.075 \\
FP8 Rollout P3O Iter 15  & 0.158 & 0.238 & \textbf{0.179} & 0.024 & 0.478 \\
FP8 Rollout P3O Iter 30  & \textbf{0.173} & \textbf{0.254} & 0.175 & \textbf{0.026} & \textbf{0.529} \\
\bottomrule
\end{tabular}
\vspace{6pt}
\caption{\textbf{Benchmark results (pass@1) across trained checkpoints.} Each group starts with its untrained model. Clip variants: Qwen3-4B-Thinking-2507 checkpoints from the clip study at iteration 35, evaluated with a 4,096-token budget; GRPO shows the mean $\pm$ population standard deviation across $\epsilon \in \{0.2, 0.4, 0.6\}$ (per-clip values at a 16,384-token budget are in \cref{tab:thinking_16k}). FP8 variants: Qwen3-8B-Base trained in BF16 with FP8 rollouts and evaluated with a 16,384-token budget; GRPO uses $\epsilon=0.4$; DAPO uses $(\epsilon_\ell, \epsilon_h)=(0.4, 0.48)$ and a different learning-rate schedule (\cref{app:dapo}). Each row other than the GRPO average is a single run, and entries are rounded half up. \textbf{Bold}: best trained checkpoint within each group per column.}
\label{tab:benchmark_results}
\end{table}

\subsection{Clip study on Qwen3-4B-Base}
\label{app:base_study}

\Cref{tab:base_benchmark} reports a clip study on Qwen3-4B-Base, evaluated on the same benchmarks with a 16,384-token budget. The runs use 16,384-token rollouts at temperature 1.0, four completions per prompt, 512 requested prompts per iteration, 50 configured iterations, seed 42 and the optimizer settings of \cref{tab:optimizer_params}. Each rollout batch receives one optimizer step. The runs differ in their GPU split, which changes the number of completions per iteration. GRPO with $\epsilon=0.2$ and $\epsilon=0.4$ uses six training and two rollout GPUs and generates $512 \times 4 = 2{,}048$ completions per iteration. GRPO with $\epsilon=0.6$ and P3O use two training and six rollout GPUs; each rollout engine takes 64 prompts, so the 512 requested prompts are rounded up to two full batches of 384 prompts, which gives 3,072 completions per iteration. In the logged configurations, the $\epsilon=0.2$ and $\epsilon=0.4$ runs differ only in the clip, and the $\epsilon=0.6$ and P3O runs differ only in the objective. The evaluated checkpoints are iterations 50, 40, 50 and 50 for $\epsilon=0.2$, $0.4$, $0.6$ and P3O, at which point the runs have generated 102,400, 81,920, 153,600 and 153,600 completions. For each run we evaluate the saved checkpoint that we judged strongest. The training prompts contain no system prompt, while the benchmark prompts contain one. The first row of \cref{tab:base_benchmark} evaluates the untrained Qwen3-4B-Base with the same settings. P3O lies between the best and the worst GRPO clip on every benchmark but is the best on none.

\begin{table}[t]
\centering
\footnotesize
\setlength{\tabcolsep}{4pt}
\renewcommand{\arraystretch}{1.2}
\begin{tabular}{lcccccc}
\toprule
\textbf{Training Method} & \textbf{Iter.} & \textbf{AIME24} & \textbf{AIME25} & \textbf{AIME26} & \textbf{AMO-Bench} & \textbf{AMC} \\
\midrule
Untrained model & -- & 0.031 & 0.058 & 0.031 & 0.013 & 0.119 \\
\midrule
GRPO ($\epsilon=0.2$) & 50 & 0.071 & 0.088 & 0.029 & 0.010 & 0.271 \\
GRPO ($\epsilon=0.4$) & 40 & \textbf{0.108} & 0.129 & 0.113 & 0.008 & \textbf{0.381} \\
GRPO ($\epsilon=0.6$) & 50 & 0.104 & 0.125 & 0.100 & 0.008 & 0.365 \\
P3O & 50 & 0.088 & 0.125 & 0.085 & 0.009 & 0.354 \\
\midrule
DAPO ($\epsilon_\ell=0.2$, $\epsilon_h=0.28$) & 30 & 0.106 & \textbf{0.175} & \textbf{0.115} & \textbf{0.013} & 0.347 \\
\bottomrule
\end{tabular}
\vspace{6pt}
\caption{\textbf{Clip study on Qwen3-4B-Base: pass@1 at a 16,384-token budget.} Iter.\ is the evaluated checkpoint. GRPO with $\epsilon=0.6$ and P3O differ only in the objective, while GRPO with $\epsilon=0.2$ and $0.4$ use a different GPU split and fewer completions per iteration (\cref{app:base_study}). The DAPO run differs from the other runs in GPU split, batch size, schedule, prompt format and rollout engine, so it is not a matched comparison (\cref{app:dapo}). Each row is a single run, and entries are rounded half up. \textbf{Bold}: best trained run per column.}
\label{tab:base_benchmark}
\end{table}

\subsection{DAPO baselines}
\label{app:dapo}

\Cref{tab:base_benchmark,tab:benchmark_results} include DAPO~\citep{yu2025dapo} as an additional baseline. The Qwen3-4B-Base run uses the clip pair $(\epsilon_\ell, \epsilon_h)=(0.2, 0.28)$ and is evaluated at iteration 30. The Qwen3-8B-Base run with FP8 rollouts uses $(0.4, 0.48)$ and is evaluated at iterations 15 and 30. Both runs use dynamic sampling, which drops prompt groups whose completions all receive the same reward and, unlike the original DAPO, does not sample new groups to replace them. They also use an overlong penalty with a 3,277-token buffer and a factor of 0.5; the benchmark evaluation applies no penalty. Both use five training and three rollout GPUs, 16,384-token rollouts, four completions per prompt and 256 requested prompts per iteration. With 16 prompts per rollout engine, the request is rounded up to 288 prompts, so each iteration generates 1,152 completions, as in the FP8 GRPO and P3O runs (\cref{tab:training_params}). At the last logged iteration, dynamic sampling kept 432 of these completions in the Base run and 240 in the FP8 run.

The FP8 run uses a different learning-rate schedule from the FP8 GRPO and P3O runs. It configures 45 iterations with 4 warmup iterations, whereas the GRPO and P3O runs configure 500 iterations with 50 warmup iterations; their logs end at iterations 45 and 35, respectively, and \cref{fig:fp8_training}a shows the first 35. At the evaluated iterations 15 and 30, the GRPO and P3O learning rates are therefore still rising ($6.9\times10^{-6}$ at iteration 15 and $8.7\times10^{-6}$ at iteration 30 in both runs, against a peak of $10^{-5}$), while the DAPO run is past its warmup. The Base run differs from the GRPO and P3O runs of \cref{app:base_study} in its GPU split, its batch (1,152 instead of 2,048 or 3,072 completions per iteration) and its schedule (40 configured iterations, and about 34,560 completions at the evaluated iteration 30). It also differs in its training prompts, which contain the system prompt used at evaluation, and in its rollout engine, which does not run in batch-invariant mode (a vLLM setting in which the output for a request does not depend on how requests are batched). Its row in \cref{tab:base_benchmark} is therefore not a matched comparison.

\subsection{Qwen3-4B-Thinking-2507 checkpoints at a 16,384-token budget}
\label{app:thinking16k}

\Cref{tab:thinking_16k} evaluates the Qwen3-4B-Thinking-2507 checkpoints of \cref{tab:benchmark_results} with a 16,384-token budget. At this budget every trained checkpoint scores below the untrained model on every benchmark. GRPO with $\epsilon=0.6$ is the best trained run on all five benchmarks, and P3O is second on four of them. The response lengths differ sharply between the runs. For GRPO with $\epsilon=0.4$ on all five benchmarks and for P3O on the three AIME sets, the mean response length equals the budget at both 4,096 and 16,384 tokens, so every one of these completions runs to the budget; the verifier still scores these truncated completions. P3O stays just below the budget on AMO-Bench and AMC (4,011 and 4,047 tokens at 4K; 16,028 and 16,080 at 16K), and GRPO with $\epsilon=0.2$ averages 15,932 to 16,294 tokens at 16K. GRPO with $\epsilon=0.6$ stops early, with means of 1,378 to 2,215 tokens at both budgets, and the untrained model averages 11,005 to 16,162 tokens at 16K.

The evaluations in \cref{tab:thinking_16k} come from two rounds that used different revisions of the evaluation code. The untrained model, GRPO with $\epsilon=0.4$, the AIME25 entry of GRPO with $\epsilon=0.6$ and the three AIME entries of P3O come from the first round, and the remaining entries from the second; the 4K-token entries of \cref{tab:benchmark_results} all come from the first round. At this budget the $\epsilon=0.6$ checkpoint path gives 73 correct AIME25 completions out of 240 in the first round and 58 in the second, and we do not know the cause; \cref{tab:thinking_16k} reports the first-round value. On the other four benchmarks of this checkpoint, the first-round 4K-token and second-round 16K-token evaluations differ by at most 9 correct completions, and its responses are short at both budgets, so we see no systematic difference between the two rounds there.

\begin{table}[t]
\centering
\footnotesize
\setlength{\tabcolsep}{4pt}
\renewcommand{\arraystretch}{1.2}
\begin{tabular}{lccccc}
\toprule
\textbf{Training Method} & \textbf{AIME24} & \textbf{AIME25} & \textbf{AIME26} & \textbf{AMO-Bench} & \textbf{AMC} \\
\midrule
Untrained model & 0.371 & 0.471 & 0.396 & 0.019 & 0.618 \\
\midrule
GRPO ($\epsilon=0.2$) & 0.033 & 0.008 & 0.017 & 0.001 & 0.103 \\
GRPO ($\epsilon=0.4$) & 0.138 & 0.171 & 0.154 & 0.014 & 0.461 \\
GRPO ($\epsilon=0.6$) & \textbf{0.202} & \textbf{0.304}$^{\dagger}$ & \textbf{0.235} & \textbf{0.018} & \textbf{0.565} \\
P3O & 0.165 & 0.183 & 0.163 & 0.009 & 0.503 \\
\bottomrule
\end{tabular}
\vspace{6pt}
\caption{\textbf{Clip study on Qwen3-4B-Thinking-2507: pass@1 at a 16,384-token budget.} The checkpoints are those of \cref{tab:benchmark_results} (iteration 35), trained with 1,024-token rollouts. $^{\dagger}$A second evaluation of the same checkpoint path, run later with a newer revision of the evaluation code, gives 0.242 (58 instead of 73 correct completions out of 240; \cref{app:thinking16k}). Each row is a single run, and entries are rounded half up. \textbf{Bold}: best trained run per column.}
\label{tab:thinking_16k}
\end{table}

\section{Additional off-policy experiments}
\label{app:offpolicy}

\subsection{Asynchronous training}
\label{app:async}

In asynchronous training, the one-step and two-step pipeline variants show the same ordering: GRPO is ahead in the middle of training and P3O at the end. The two variants use the same maximum lag of 10 policy versions and differ only in the number of optimizer steps per iteration. We show the one-step setting in \cref{fig:async_training}; its caption also gives the two-step results.

\begin{figure}[t]
    \centering
    {\small%
    \tikz[baseline=-0.6ex]{\draw[cb-red,line width=2.2pt](0,0)--(12pt,0);}\;P3O%
    \hspace{1.6em}%
    \tikz[baseline=-0.6ex]{\draw[cb-blue,line width=1.6pt](0,0)--(12pt,0);}\;GRPO%
    }

    \begin{tikzpicture}
        \begin{axis}[
            width=0.65\textwidth,
            height=0.4\textwidth,
            xlabel={\textbf{Training Steps}},
            ylabel={\textbf{Average Reward}},
            xlabel style={font=\small},
            ylabel style={font=\small},
            grid=major,
            grid style={dashed, gray!30},
            axis x line=bottom,
            axis y line=left,
            enlarge x limits=false,
            ymin=0,
            ymax=0.8,
    ]

    \addplot[color=cb-blue, line width=1.2pt] table[x=step, y=grpo_async_m5, col sep=comma] {plot_data/async_plot.csv};
    \addplot[color=cb-red, line width=1.8pt] table[x=step, y=p3o_async_m5, col sep=comma] {plot_data/async_plot.csv};

    \end{axis}
    \end{tikzpicture}
    \caption{\textbf{Asynchronous-training comparison between P3O and GRPO under one optimizer step per iteration.}
    Qwen3-4B-Thinking-2507 is trained on GSM8K~\citep{cobbe2021gsm8k} with 1,024-token rollouts, and the
    replay buffer keeps rollouts that are at most 10 policy versions old.
    Rollouts are generated by a stale policy while the training engine continues updating. GRPO uses the decoupled loss~\citep{hilton2021batch,fu2025areal} with
    $\epsilon=0.4$ and $c_w=5$, while P3O uses \cref{eq:p3o_intro} unchanged. GRPO takes its proximal snapshot before the single
    optimizer step of each iteration, so $\pi_{\mathrm{prox}}=\pi_\theta$ at that step, the $\epsilon$ clip does not bind, and the GRPO gradient
    equals that of the CISPO row of \cref{tab:objective_comparison} with the cap $1+\epsilon_h$ set to $c_w=5$. In this one-step pipeline
    setting, both methods peak at step 3 (0.651 for P3O and 0.665 for GRPO). GRPO then drops to 0.237 at step 6
    and P3O to 0.438 at step 7, and GRPO is ahead at 10 of the 11 steps from
    10 to 20. P3O is ahead at 22 of the last 23 steps and ends at 0.607 against 0.510. With two
    optimizer steps per iteration (not shown), P3O is ahead from iteration 37 on and ends at 0.697
    against 0.618.}
    \label{fig:async_training}
\end{figure}

\subsection{Sampling temperature on Qwen2.5-1.5B-Instruct}
\label{app:temp_qwen25}

\Cref{fig:temp_variability_appendix} shows the Qwen2.5-1.5B-Instruct runs at $T=0.6$ and $T=1.2$: at $T=0.6$ both methods collapse, and at $T=1.2$ neither degrades.

\begin{figure}[t]
    \centering
    {\small%
    \tikz[baseline=-0.6ex]{\draw[cb-red,line width=2.2pt](0,0)--(12pt,0);}\;P3O%
    \hspace{1.6em}%
    \tikz[baseline=-0.6ex]{\draw[cb-blue,line width=1.6pt](0,0)--(12pt,0);}\;GRPO%
    }

    \begin{subfigure}[b]{0.49\textwidth}
        \centering
        \begin{tikzpicture}
    \begin{axis}[
        width=\textwidth,
        height=0.6\textwidth,
        xlabel={\textbf{Training Steps}},
        xlabel style={font=\small},
        ylabel={\textbf{Average Reward}},
        ylabel style={font=\small},
        grid=major,
        grid style={dashed, gray!30},
        axis x line=bottom,
        axis y line=left,
        enlarge x limits=false,
        ymin=0,
        ymax=0.35,
    ]

    \addplot[color=cb-blue, line width=1.2pt] table[x=step, y=grpo, col sep=comma] {plot_data/temp_0.6_plot_qwen25.csv};
    \addplot[color=cb-red, line width=1.8pt] table[x=step, y=p3o, col sep=comma] {plot_data/temp_0.6_plot_qwen25.csv};

    \end{axis}
\end{tikzpicture}
        \caption{Qwen2.5-1.5B-Instruct (Temperature 0.6)}
    \end{subfigure}
    \begin{subfigure}[b]{0.49\textwidth}
        \centering
        \begin{tikzpicture}
    \begin{axis}[
        width=\textwidth,
        height=0.6\textwidth,
        xlabel={\textbf{Training Steps}},
        xlabel style={font=\small},
        grid=major,
        grid style={dashed, gray!30},
        axis x line=bottom,
        axis y line=left,
        enlarge x limits=false,
        ymin=0,
        ymax=0.35,
    ]

    \addplot[color=cb-blue, line width=1.2pt] table[x=step, y=grpo, col sep=comma] {plot_data/temp_1.2_plot_qwen25.csv};
    \addplot[color=cb-red, line width=1.8pt] table[x=step, y=p3o, col sep=comma] {plot_data/temp_1.2_plot_qwen25.csv};

    \end{axis}
\end{tikzpicture}
        \caption{Qwen2.5-1.5B-Instruct (Temperature 1.2)}
    \end{subfigure}
    \caption{\textbf{Temperature results for Qwen2.5-1.5B-Instruct, corresponding to \cref{fig:temp_variability}.} At $T=0.6$ both methods collapse (mean reward over the last five steps of 0.001 for P3O and 0.013 for GRPO), and at $T=1.2$ neither degrades.}
    \label{fig:temp_variability_appendix}
\end{figure}

\section{Two-anchor extension of P3O}
\label{app:two_anchor}

\subsection{Full formulation}
\label{app:p3o_extension}

One possible issue with P3O is that the single ESS $e_{\mathcal{B}}$ in~\cref{eq:p3o_intro} conflates two distinct drifts in the batch: the difference between $\pi_\theta$ and the behavior policy $\pi_{\mathrm{b}}$, and the within-iteration difference between $\pi_\theta$ and its pre-update snapshot $\pi_{\mathrm{prox}}$. When both drifts are large but in different directions, a single anchor cannot distinguish them. A natural extension replaces the single-anchor regularizer of~\cref{eq:p3o_intro} with a KL toward an ESS-weighted mixture of the behavior policy $\pi_{\mathrm{b}}$ and a proximal snapshot $\pi_{\mathrm{prox}}$ of $\pi_\theta$ taken at the start of each iteration. The full loss is
\begin{align}
    \mathcal{L}_{\mathrm{ext}}(\theta) = -\E_{\pi_{\mathrm{b}}}\!\bigl[\sg(\min\{r_b,\,e_{\mathrm{mix}}\})\,\log\pi_\theta(y_t \mid c_{<t})\,A\bigr] + (1-e_{\mathrm{mix}})\,\KL\!\bigl(\pi_\theta\,\big\|\,\pi_{\mathrm{mix}}\bigr),
    \label{eq:two_anchor_loss_appendix}
\end{align}
with the following per-batch quantities, where $r_b$ is the ratio $\rho_t$ of~\cref{eq:token_ratio} and $e_b$ equals $e_{\mathcal{B}}$ (\cref{eq:ess_def}):
\begin{align}
    r_b &= \frac{\pi_\theta(y_t\mid c_{<t})}{\pi_{\mathrm{b}}(y_t\mid c_{<t})}, \quad
    r_{\mathrm{prox}} = \frac{\pi_\theta(y_t\mid c_{<t})}{\pi_{\mathrm{prox}}(y_t\mid c_{<t})}, \nonumber \\
    e_b &= \frac{\bigl(\sum_\mathcal{B} r_b\bigr)^2}{|\mathcal{B}|\,\sum_\mathcal{B} r_b^{2}}, \quad
    e_{\mathrm{prox}} = \frac{\bigl(\sum_\mathcal{B} r_{\mathrm{prox}}\bigr)^2}{|\mathcal{B}|\,\sum_\mathcal{B} r_{\mathrm{prox}}^{2}}, \nonumber \\
    e_{\mathrm{mix}} &= \min(e_b,\,e_{\mathrm{prox}}), \nonumber \\
    \pi_{\mathrm{mix}}(\cdot \mid c_{<t}) &= \frac{(1-e_b)\,\pi_{\mathrm{b}}(\cdot \mid c_{<t}) + (1-e_{\mathrm{prox}})\,\pi_{\mathrm{prox}}(\cdot \mid c_{<t})}{(1-e_b) + (1-e_{\mathrm{prox}})}.
    \label{eq:two_anchor_defs_appendix}
\end{align}
As in P3O, $e_b$ and $e_{\mathrm{prox}}$ (and thus $e_{\mathrm{mix}}$ and the two mixture weights) are detached, and $\pi_{\mathrm{prox}}$ receives no gradient. When $1-e_{\mathrm{mix}}\le10^{-6}$, which includes the case $e_b=e_{\mathrm{prox}}=1$ where $\pi_{\mathrm{mix}}$ is undefined, we set the regularizer to zero. Note that both ESS values are computed on the tokens of $\mathcal{B}$, which are sampled from $\pi_{\mathrm{b}}$. Thus $e_{\mathrm{prox}}$ is not the importance-sampling ESS of samples from $\pi_{\mathrm{prox}}$; it measures the relative spread of the ratios $r_{\mathrm{prox}}$ over $\mathcal{B}$ and equals one when they are all equal (e.g., when $\pi_\theta=\pi_{\mathrm{prox}}$).
Each anchor is weighted by its $(1-\mathrm{ESS})$, so a fully on-policy axis (whose ESS approaches one) drops out of the mixture and the regularizer pulls only toward the mismatched anchor. The construction reduces to P3O when $e_{\mathrm{prox}}\to1$: the proximal weight then vanishes, $\pi_{\mathrm{mix}}\to\pi_{\mathrm{b}}$ and $e_{\mathrm{mix}}\to e_b$, recovering~\cref{eq:p3o_intro}. If instead $e_b\to1$, the behavior weight vanishes and $\pi_{\mathrm{mix}}\to\pi_{\mathrm{prox}}$, so the regularizer anchors at the proximal snapshot and not at $\pi_{\mathrm{b}}$. The construction adds no hyper-parameter beyond what P3O already uses. The second anchor has an effect only when an iteration takes more than one optimizer step: with one step, as in the clip and FP8 runs (\cref{tab:training_params}), $\pi_{\mathrm{prox}}=\pi_\theta$ at that step, so $e_{\mathrm{prox}}=1$ and the extension reduces to P3O.

\subsection{Training stability}
\label{app:two_anchor_results}

\Cref{fig:two_anchor_reward} reports training curves for the two-anchor extension of P3O described
in~\cref{app:p3o_extension}, alongside P3O and GRPO baselines.
All three algorithms are trained on Qwen3-4B-Thinking-2507 using the DeepScaleR-Preview dataset with
temperature~1.2 and a rollout length of 4,096~tokens. The number of optimizer steps per iteration of the two-anchor runs was not recorded.

The two-anchor extension comes close to the peak reward of P3O and GRPO (0.673 against 0.694 and 0.697) during the first 16 steps but then
collapses abruptly, falling to near-zero reward (0.012) before the run terminates at step~24.
P3O remains stable throughout the full training run, and GRPO collapses later, falling from 0.567 at step~21 to 0.199 at step~23 (\cref{fig:temp_variability}).
We have not identified the cause of the collapse. Addressing this instability (e.g., through tighter proximal resets, adaptive KL
weighting, or learning-rate schedules) is left as future work.

\begin{figure}[t]
    \centering
    {\small%
    \tikz[baseline=-0.6ex]{\draw[cb-red,line width=2.2pt](0,0)--(12pt,0);}\;P3O%
    \hspace{1.6em}%
    \tikz[baseline=-0.6ex]{\draw[cb-blue,line width=1.6pt](0,0)--(12pt,0);}\;GRPO%
    \hspace{1.6em}%
    \tikz[baseline=-0.6ex]{\draw[cb-green,line width=1.6pt](0,0)--(12pt,0);}\;Two-anchor extension%
    }

    \begin{tikzpicture}
        \begin{axis}[
            width=0.65\textwidth,
            height=0.4\textwidth,
            xlabel={\textbf{Training Steps}},
            ylabel={\textbf{Average Reward}},
            xlabel style={font=\small},
            ylabel style={font=\small},
            grid=major,
            grid style={dashed, gray!30},
            axis x line=bottom,
            axis y line=left,
            enlarge x limits=false,
            ymin=0,
            ymax=0.8,
    ]

    \addplot[color=cb-green, line width=1.2pt] table[x=step, y=p4o, col sep=comma] {plot_data/p4o_reward_plot.csv};
    \addplot[color=cb-blue, line width=1.2pt] table[x=step, y=grpo, col sep=comma] {plot_data/p4o_reward_plot.csv};
    \addplot[color=cb-red, line width=1.8pt] table[x=step, y=p3o, col sep=comma] {plot_data/p4o_reward_plot.csv};

    \end{axis}
    \end{tikzpicture}
    \caption{\textbf{Training curves of the two-anchor extension, P3O, and GRPO on Qwen3-4B-Thinking-2507.}
    All runs use the DeepScaleR-Preview dataset with temperature~1.2 and a rollout length of 4,096~tokens.
    The two-anchor extension peaks at a reward comparable to P3O and GRPO but undergoes a sharp collapse after step~16,
    dropping to near-zero reward by step~24 before the run terminates.
    P3O stays between 0.56 and 0.70 from step~10 to step~36, while GRPO collapses later, falling from 0.567 at step~21 to 0.199 at step~23.}
    \label{fig:two_anchor_reward}
\end{figure}

\Cref{fig:two_anchor_reward2} presents a complementary experiment at default rollout temperature with 1,024-token rollouts (its P3O and GRPO curves are the clip runs of~\cref{tab:training_params}), where the same three methods are compared and GRPO again uses $\epsilon{=}0.4$.

\begin{figure}[t]
    \centering
    {\small%
    \tikz[baseline=-0.6ex]{\draw[cb-red,line width=2.2pt](0,0)--(12pt,0);}\;P3O%
    \hspace{1.6em}%
    \tikz[baseline=-0.6ex]{\draw[cb-blue,line width=1.6pt](0,0)--(12pt,0);}\;GRPO, $\epsilon=0.4$%
    \hspace{1.6em}%
    \tikz[baseline=-0.6ex]{\draw[cb-green,line width=1.6pt](0,0)--(12pt,0);}\;Two-anchor extension%
    }

    \begin{tikzpicture}
        \begin{axis}[
            width=0.65\textwidth,
            height=0.4\textwidth,
            xlabel={\textbf{Training Steps}},
            ylabel={\textbf{Average Reward}},
            xlabel style={font=\small},
            ylabel style={font=\small},
            grid=major,
            grid style={dashed, gray!30},
            axis x line=bottom,
            axis y line=left,
            enlarge x limits=false,
            ymin=0,
            ymax=0.6,
    ]

    \addplot[color=cb-green, line width=1.2pt] table[x=step, y=p4o, col sep=comma] {plot_data/p4o_reward_plot2.csv};
    \addplot[color=cb-blue, line width=1.2pt] table[x=step, y=grpo_clip04, col sep=comma] {plot_data/p4o_reward_plot2.csv};
    \addplot[color=cb-red, line width=1.8pt] table[x=step, y=p3o, col sep=comma] {plot_data/p4o_reward_plot2.csv};

    \end{axis}
    \end{tikzpicture}
    \caption{\textbf{Training curves of the two-anchor extension, P3O, and GRPO ($\epsilon{=}0.4$) on Qwen3-4B-Thinking-2507
    at default temperature.}
    All runs use the DeepScaleR-Preview dataset with a rollout length of 1,024~tokens.
    Unlike the temperature-1.2 regime (\cref{fig:two_anchor_reward}), the two-anchor extension remains stable throughout
    training and ends at 0.497, against 0.508 for P3O and 0.438 for GRPO with $\epsilon{=}0.4$;
    over the last five steps it averages 0.515 against 0.510 for P3O.}
    \label{fig:two_anchor_reward2}
\end{figure}

\section{Additional related work}
\label{app:related_work}

\paragraph{Quantized rollouts.} QuRL~\citep{li2026qurl} generates rollouts with a quantized (INT8 or FP8) actor. To prevent long-term collapse, it sets the clip range adaptively from the ratio between the full-precision and the quantized actor, and it rescales the weights (invariant scaling) so that the small weight changes between RL steps are not lost to quantization. Its finding that a fixed clip range is not enough under quantized rollouts is consistent with our FP8 result on Qwen3-8B-Base (\cref{fig:fp8_training}a). The two methods differ in how they respond: QuRL makes the clip adaptive for this source of mismatch, whereas P3O removes the clip and sets its cap and regularizer from the batch ESS, with the same objective in every setting we test. Invariant scaling acts on the quantized weights and not on the objective, so it could be combined with P3O; we have not tested this combination. QUADS~\citep{zhuge2026quads} studies NVFP4 rollouts of mixture-of-experts models trained in BF16 with GRPO and aligns the quantization error on both sides: the trainer fake-quantizes the weights, and the rollout engine compensates the activation channels with the largest error. QUADS leaves the GRPO objective unchanged, so it is complementary to P3O as well.

\paragraph{In-flight weight updates.} PipelineRL~\citep{pipelinerl} runs generation and training concurrently and updates the weights of the rollout engines after each optimizer step without restarting the sequences in progress, so the tokens of one sequence can come from several policy versions. Since P3O takes each ratio per token against the log-probability that the rollout engine recorded (\cref{sec:approach}), such sequences need no single behavior policy. We have not tested P3O with in-flight weight updates.

\section{Additional experimental details}
\label{app:exp_details}

This section summarizes the settings for the main runs.

\begin{table}[!ht]
\centering
\footnotesize
\setlength{\tabcolsep}{4pt}
\begin{tabular}{|l|c|}
\hline
\textbf{Hyper-parameter} & \textbf{Value} \\ \hline
Optimizer & AdamW \\ \hline
Learning rate & 1e-5 \\ \hline
Betas ($\beta_1, \beta_2$) & (0.9, 0.95) \\ \hline
Weight decay & 0.01 \\ \hline
Gradient clipping & 1.0 \\ \hline
LR scheduler & WarmupCosineLR (10\% warmup ratio) \\ \hline
\end{tabular}
\vspace{4pt}
\caption{\textbf{Shared optimizer settings used across all experiments.} The warmup covers 10\% of the configured iterations of each run, rounded down. The FP8 GRPO and P3O runs configure 500 iterations, so their checkpoints at iterations 15 and 30 are within the 50-iteration warmup (\cref{app:dapo}).}
\label{tab:optimizer_params}
\end{table}

\begin{table}[!ht]
\centering
\footnotesize
\setlength{\tabcolsep}{4pt}
\begin{tabular}{|l|c|c|c|}
\hline
\textbf{Hyper-parameter} & \textbf{Clip Runs} & \textbf{Temperature Runs} & \textbf{FP8 Runs} \\ \hline
Model & Qwen3-4B-Thinking-2507 & Qwen3-4B-Thinking-2507 & Qwen3-8B-Base \\ \hline
Training / rollout GPU split & 6 train + 2 rollout & 6 train + 2 rollout & 5 train + 3 rollout \\ \hline
Optimizer steps / iteration & 1 & -- & 1 \\ \hline
Train micro-batch / GPU & 8 & 4 & 1 \\ \hline
Gradient accumulation & 1 & 2 & 1 \\ \hline
Rollout batch size / GPU & 64 & 64 & 16 \\ \hline
Requested prompts / iteration & 512 & 512 & 256 \\ \hline
Completions / iteration & 2{,}048 & 2{,}048 & 1{,}152 \\ \hline
Max tokens & 1{,}024 & 4{,}096 & 16{,}384 \\ \hline
Sampling temperature & 1.0 & $T \in \{0.6, 1.2\}$ & 1.0 \\ \hline
GRPO clip setting & $\epsilon \in \{0.2, 0.4, 0.6\}$ & $\epsilon = 0.4$ & $\epsilon = 0.4$ \\ \hline
\end{tabular}
\vspace{4pt}
\caption{\textbf{Per-experiment training configuration.} Completions per iteration are the requested prompts times four completions per prompt, after the rollout engines round the prompt count up to full batches (256 requested prompts become 288 with three engines of 16 prompts each). In the clip and FP8 runs, each iteration takes one optimizer step on the gradients accumulated over all its micro-batches, so the gradient-accumulation row (a DeepSpeed setting) does not change the update. The micro-batch row, together with the number of training GPUs, sets the tokens over which P3O computes each ESS (\cref{eq:ess_def}). The configuration files of the temperature runs were not archived; their column gives the settings from our run notes, which do not record the number of optimizer steps per iteration. The Qwen2.5-1.5B-Instruct GRPO clip runs use the settings of the clip column other than the model; the configurations of the other Qwen2.5-1.5B-Instruct runs were not archived.}
\label{tab:training_params}
\end{table}

\begin{table}[!ht]
\centering
\renewcommand{\arraystretch}{0.92}
\footnotesize
\begin{tabular}{|l|c|}
\hline
\textbf{Hyper-parameter} & \textbf{Value} \\ \hline
\multicolumn{2}{|c|}{\textit{Shared benchmark-evaluation settings}} \\ \hline
Temperature & 1.0 \\
Top-$p$ & 0.95 \\
Samples / problem & 16 \\
Rollout Batch Size / GPU & 16 \\
Rollout GPUs & 8 \\
Data Workers & 8 \\
Benchmarks (problems) & AIME24 (30), AIME25 (15), AIME26 (30), AMO-Bench (50), AMC (45) \\ \hline
\multicolumn{2}{|c|}{\textit{4K-token budget (Qwen3-4B-Thinking-2507 clip study)}} \\ \hline
Max Response Tokens & 4{,}096 \\ \hline
\multicolumn{2}{|c|}{\textit{16K-token budget (FP8, Qwen3-4B-Base and Qwen3-4B-Thinking-2507 checkpoints)}} \\ \hline
Max Response Tokens & 16{,}384 \\ \hline
\multicolumn{2}{|c|}{\textit{Model \& system configuration}} \\ \hline
Precision & bfloat16 \\
Tensor Parallel Size & 1 \\
\hline
\end{tabular}
\vspace{4pt}
\caption{\textbf{Benchmark-evaluation settings, grouped by evaluation budget.} We evaluate with a 4K-token or a 16K-token budget; all other sampling settings and the GPU allocation are the same for every benchmark and checkpoint.}
\label{tab:evaluation_params}
\end{table}

\begin{table}[!ht]
\centering
\renewcommand{\arraystretch}{0.92}
\footnotesize
\begin{tabular}{|l|c|}
\hline
\textbf{Setting} & \textbf{Value} \\ \hline
\multicolumn{2}{|c|}{\textit{Accelerator Configuration}} \\ \hline
GPU Model & NVIDIA H100 (clip, temperature and FP8 runs) \\
 & NVIDIA A100 80GB (\cref{fig:async_training}) \\
Total GPUs per run & 8 \\
Evaluation partition & 8 rollout GPUs \\
Tensor Parallel Size & 1 \\ \hline
\multicolumn{2}{|c|}{\textit{Runtime Configuration}} \\ \hline
Training Precision & bfloat16 \\
Rollout Precision & bfloat16; FP8 only in the FP8 runs \\
Rollout Engine Mode & vLLM batch-invariant where logged (off in FP8 and DAPO runs) \\
Distributed Training & DeepSpeed ZeRO-3 \\
Trainer Attention Backend & Flash Attention 2 \\
\hline
\multicolumn{2}{|c|}{\textit{Workload Summary}} \\ \hline
Training Dataset & DeepScaleR-Preview; GSM8K for \cref{fig:async_training} \\
Training Models & Qwen3-4B-Thinking-2507, Qwen3-4B-Base, \\
 & Qwen3-8B-Base, Qwen2.5-1.5B-Instruct \\
Evaluation Rollout Batch Size / GPU & 16 \\ \hline
\multicolumn{2}{|c|}{\textit{Software Environment}} \\ \hline
Python version & 3.13.1 or 3.12 (recommended per code revision) \\
NVIDIA driver & CUDA 12.x-compatible; driver version $\geq$ 525.85 recommended \\
CUDA toolkit & 12.2 (tested); toolkit versions $\geq$ 12.2 supported \\
PyTorch build & CUDA 12.6 wheels \\ \hline
\multicolumn{2}{|c|}{\textit{Training and Rollout Stack}} \\ \hline
PyTorch / TorchVision & CUDA-enabled install \\
DeepSpeed & 0.18.9 \\
vLLM & 0.19.0 \\
Transformers & 4.57.6 \\
Ray & 2.54.1 \\
FlashAttention & 2.8.3 (built from source) \\ \hline
\end{tabular}
\vspace{4pt}
\caption{\textbf{Compute layout and software environment for the reported runs.} The eight GPUs are partitioned differently across runs (\cref{tab:training_params}; \cref{app:base_study,app:dapo}). The software versions are those that the training code pins or recommends at the revisions used for the runs; the runtime versions were not logged.}
\label{tab:hardware_config}
\end{table}

\Cref{tab:training_params} lists the settings of the three Qwen3 experiment families: the clip and temperature runs train Qwen3-4B-Thinking-2507 with 1,024- and 4,096-token rollouts on six training and two rollout GPUs, and the FP8 runs train Qwen3-8B-Base with 16,384-token rollouts on five training and three rollout GPUs. The benchmark prompts contain a system prompt; the FP8 runs train with it and the clip runs without it. \Cref{tab:benchmark_results} evaluates the clip-study checkpoints at 4,096 tokens and all others at 16,384 tokens; \cref{app:thinking16k} also evaluates the clip-study checkpoints at 16,384 tokens.

\paragraph{Reproducibility.} Each configuration is run once with a fixed seed for Python, NumPy, PyTorch, and rollout sampling. We further reduce nondeterminism using vLLM batch-invariant mode\footnote{\url{https://docs.vllm.ai/en/latest/features/batch_invariance/}}, deterministic cuBLAS and cuDNN settings, disabled cuDNN benchmarking, and a fixed \texttt{PYTHONHASHSEED}. Some nondeterminism remains, e.g., from distributed floating-point reductions.


\section*{AI Use Statement}
We did not use generative AI tools to develop the method or theory, design or run experiments, or generate, clean, or reformat data. We used them mainly for writing and presentation assistance, including grammar correction, improving sentence flow, clarity, and readability, polishing the text, checking references and related work, writing code to improve the formatting and visual presentation of plots, and verifying reported experimental results against the associated logs. All AI-assisted content was reviewed and verified by the authors.
\end{document}